\documentclass[review]{elsarticle}
\usepackage{lineno,hyperref}
\usepackage{float}
\usepackage{caption} 
\usepackage{multirow}
\biboptions{sort&compress}
\hypersetup{hidelinks}
\modulolinenumbers[5]
\graphicspath{ {./figure/} }

\usepackage{caption}
\usepackage{geometry} 
\usepackage{array, makecell}

\usepackage{hyperref}
\usepackage{xspace}
\newcommand*{\eg}{e.g.,\@\xspace}
\newcommand*{\ie}{i.e.,\@\xspace}

\journal{Advanced Engineering Informatics}

\begin{document}

\begin{frontmatter}

\title{Automatic Roof Type Classification Through Machine Learning for Regional Wind Risk Assessment}



\author[mymainaddress]{Shuochuan Meng\corref{correspondingauthor}}
\cortext[correspondingauthor]{Corresponding author}
\ead{vessel@ucla.edu}
\author[mymainaddress]{Mohammad Hesam Soleimani-Babakamali}
\author[mymainaddress]{Ertugrul Taciroglu}

\address[mymainaddress]{Department of Civil and Environmental Engineering, University of California, Los Angeles, CA, United States}

\begin{abstract}
Roof type is one of the most critical building characteristics for wind vulnerability modeling. It is also the most frequently missing building feature from publicly available databases. An automatic roof classification framework is developed herein to generate high-resolution roof-type data using machine learning. A Convolutional Neural Network (CNN) was trained to classify roof types using building-level satellite images. The model achieved the F1 score of 0.96 on predicting roof types for 1,000 test buildings. The CNN model was then used to predict roof types for 161,772 single-family houses in New Hanover County, NC, and Miami-Dade County, FL. The distribution of roof type in city and census tract scales was presented. A high variance was observed in the dominant roof type among census tracts. To improve the completeness of the roof-type data, imputation algorithms were developed to populate missing roof data due to low-quality images, using critical building attributes and neighborhood-level roof characteristics.  
\end{abstract}


\end{frontmatter}

\section{Introduction}

\noindent Roof type is one of the most critical building characteristics for wind vulnerability modeling of residential buildings. In hurricane-prone regions of the United States, gable and hip roofs are the dominant roof types for single-family houses~\cite{pita2008,model2005}. It was observed in wind tunnels tests~\cite{xu1998,gavanski2013,shao2018} and post-disaster survey~\cite{Crandell1993,Brown-Giammanco2018} that gable and hip roofs experience distinctive wind pressures and damage frequency under high winds. Therefore, the quality of roof-type data is crucial for the accuracy and reliability of regional wind risk assessments. Nevertheless, roof type is the also most frequently missing building feature from publicly available databases such as tax appraisers' databases~\cite{pita2008}.

Several studies have focused on populating roof-type data using machine learning-based methods, which can be summarized into two major types. One type of approach is predicting roof type using other building features~\cite{pita2011,mohajeri2018} with machine learning models. Pita et al.~\cite{pita2011} applied Bayesian Belief Networks (BBN) and Classification and Regression Trees (CART) to impute missing roof-type data using building characteristics such as construction year and building value. Mohajeri et al.~\cite{mohajeri2018} used Support Vector Machine (SVM) to classify roof types using roof geometries obtained from LiDAR data. However, the prediction of roof types heavily relied on the availability of other building information. Moreover, building features, like building value, are correlated with building location, which deteriorates the performance of the model when applied to different areas~\cite{pita2011}. The other approach is classifying roof types using remote sensing data with Convolutional Neural Networks (CNNs) through Representation Learning~\cite{bengio2013}. Representation learning extracts the features automatically through its learning process, without the need for hand-crafted features~\cite{HUTHWOHL2018150}, thus offering higher generalizability than statistical and hand-crafted features which might not generalize to unseen data well~\cite{el2019}. Various CNNs were trained to predict roof type using neighborhood-level satellites image~\cite{alidoost2018}, single-building-level satellite images~\cite{buyukdemircioglu2021,wang2021}, and the combination of satellite images and LiDAR data~\cite{castagno2018}. It was demonstrated that CNNs could classify roof types efficiently with high accuracy. Nevertheless, low-quality satellite images are frequently encountered due to, for example, occlusion by trees, which reduces the robustness of satellite-image-based classification or segmentation~\cite{jayaseeli2020,zambanini2020}, including the roof classification objective herein. Therefore, devising data imputation algorithms are one possible approach to tackle this problem.
  
Building archetypes are defined and mapped to the building inventory in existing wind vulnerability models, including the Federal Emergency Management Agency (FEMA) HAZUS-MH model~\cite{vickery2006} and the Florida Public Hurricane Loss Model (FPHLM)~\cite{pinelli2011}. Rectangular-shaped buildings with simple gable and hip roofs were used to model single-family houses for wind damage assessment. However, most single-family residential buildings have non-rectangular footprints with resulting complex roof configurations. Wind tunnel tests~\cite{shao2018,parackal2016,uematsu2022,sarma2023} showed different magnitudes and distributions of wind pressures on simple roofs and complex roofs. In addition, wind fragility analysis methodology was developed for buildings with complex roofs in recent studies~\cite{masoomi2018,stewart2018,sarma2023}. Therefore, building archetypes for single-family houses need to be refined by including more roof types to provide a more realistic description of the building stock.    
   
In this paper, five roof classes are defined to incorporate variations of roof types (gable or hip) and building shapes (rectangular or non-rectangular). An automatic roof classification framework is proposed to generate roof-type data for the use of regional wind risk assessments. First, roof types are classified using building-level satellite images with CNNs and the low-quality images are detected and removed. Next, the missing roof-type data are imputed using other building characteristics and the roof type distribution of the neighboring buildings. The proposed framework is applied to two study areas in the United States, New Hanover County, NC, and Miami-Dade County, FL. Roof type is classified for 161,772 single-family houses, and the distribution of roof types is shown at the city and census tract levels.

The main contributions of this paper are summarized below:
\begin{itemize}
\item {Proposed an automatic roof type classification framework that can generate high-resolution roof-type data for regional wind risk assessment.}
\item {Generated city-scale roof-type datasets, which revealed the spatial variance of dominant roof type at neighborhood, census tract, and city scales.}
\item {Investigated the relative importance between different building characteristics for predicting the roof type.}
\end{itemize}

\section{Roof classification workflow and case study area}\label{overview}

\begin{figure}[H]
\centering
\includegraphics[scale=0.45]{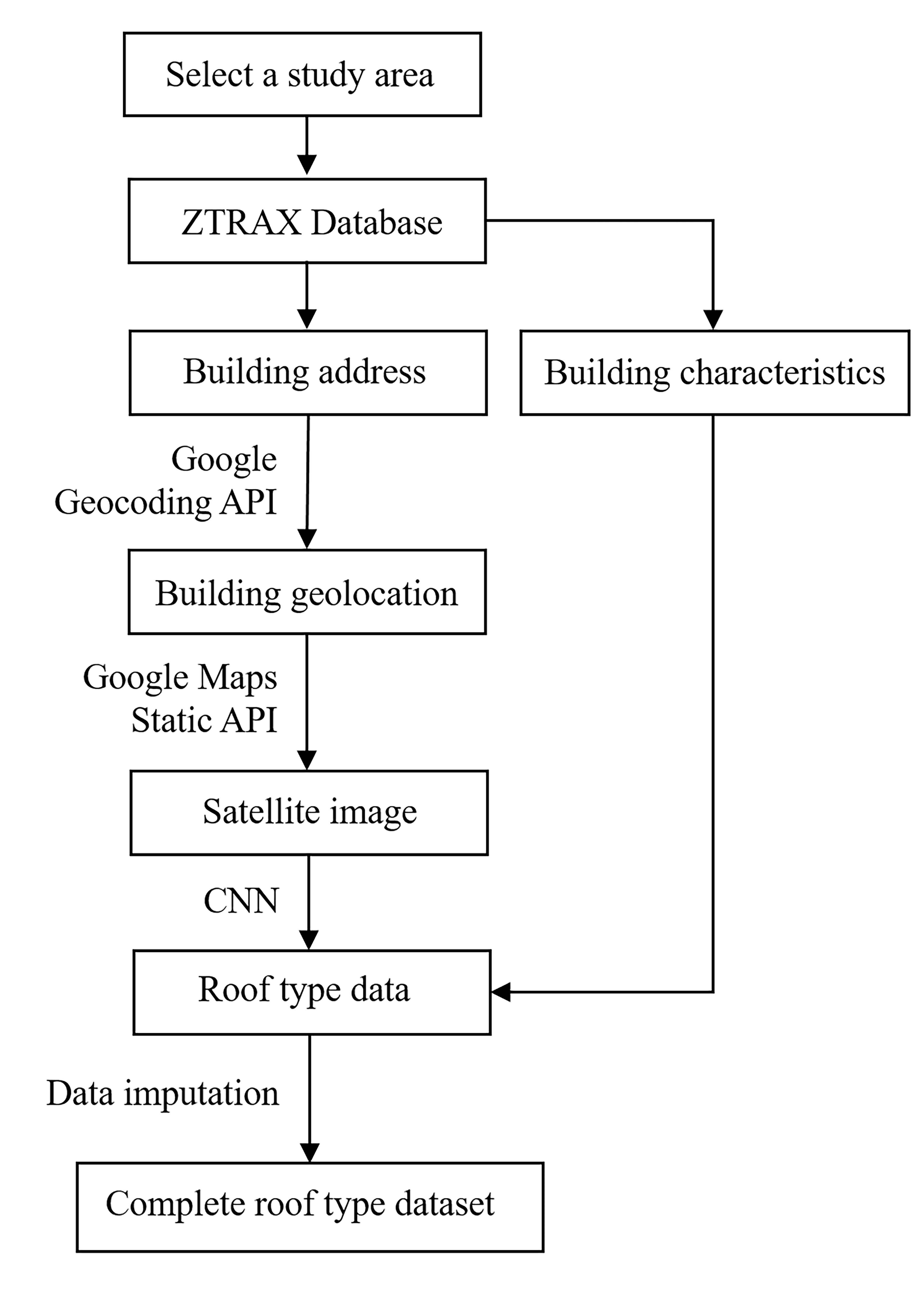}
\caption{Proposed roof type classification workflow.}
\label{flowchart}
\end{figure}

\noindent The proposed city-level roof type classification workflow for single-family houses is shown in Fig~\ref{flowchart}. First, building address and characteristics (\ie year built and building area) for single-family dwellings in a chosen area are obtained from the ZTRAX database~\cite{zillow2018}, the largest real estate database in the United States provided by Zillow. Next, building geolocation (\ie longitude and latitude of the centroid of the building) is acquired based on the building address using Google Geocoding API \footnote{\url{https://developers.google.com/maps/documentation/geocoding}}. A satellite image of each building is then downloaded using Google Maps Static API \footnote{\url{https://developers.google.com/maps/documentation/maps-static}} and cropped based on floor area to optimize the scale of the image. Following the collection of satellite images, a CNN model is applied to the images to classify roof types and identify low-quality images that do not contain recognizable roofs. In the end, the missing data resulting from the image-based classification due to low-quality images are populated by a data imputation algorithm. 

\begin{figure}[H]
\centering
\includegraphics[scale=0.55]{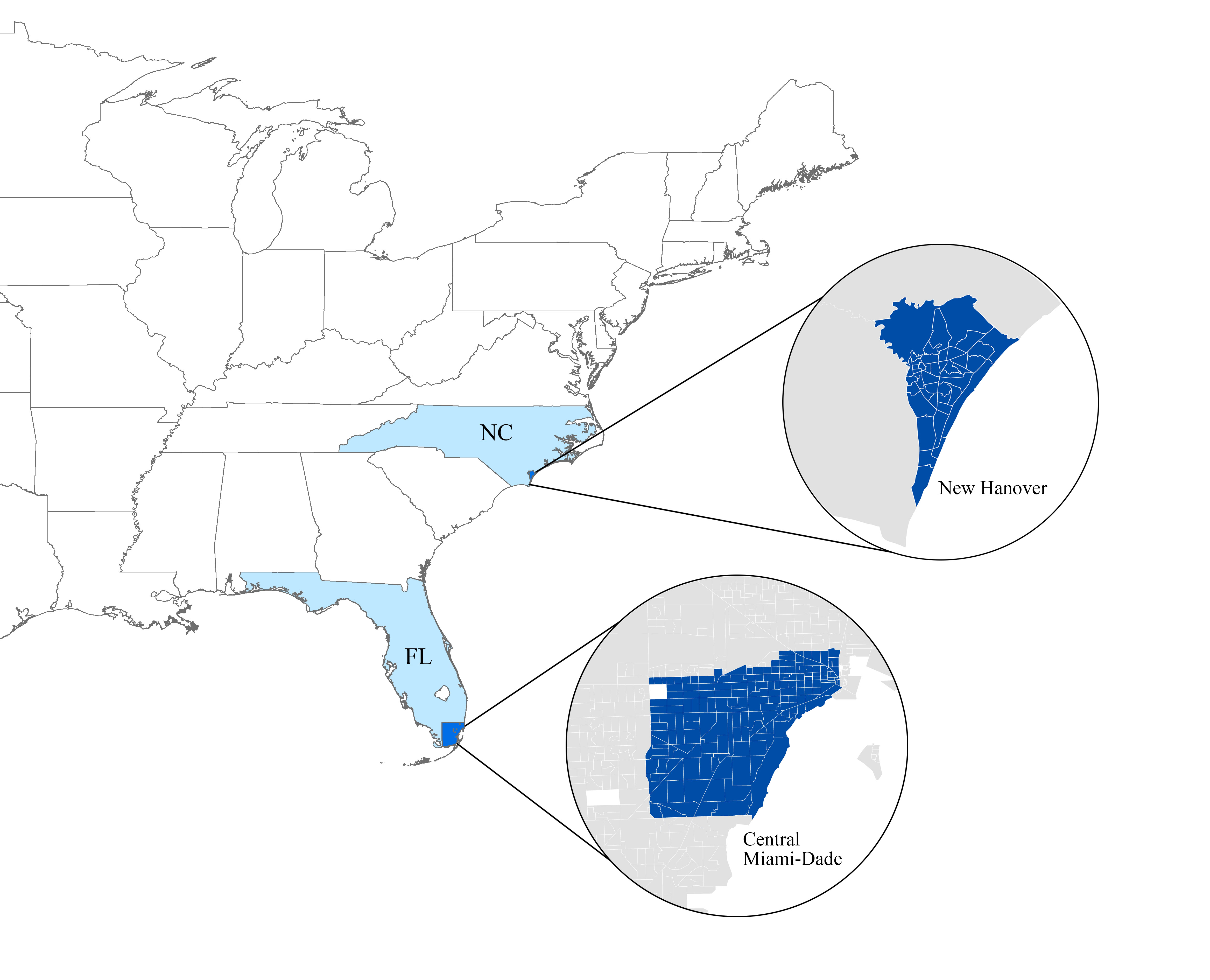}
\caption{Location of study areas}
\label{case_loc}
\end{figure}

The roof classification workflow was applied to two case study areas, New Hanover County, North Carolina, and the central region of Miami-Dade County, Florida, to generate a city-level roof-type dataset. Locations of study areas are illustrated in Fig.~\ref{case_loc}. Extensive studies on regional wind risk assessment have been conducted on the selected study areas~\cite{vickery2006,pinelli2011,Peng2013}. Nevertheless, detailed roof-type datasets are not available for these areas. The building addresses for 68,504 and 93,268 single-family houses were obtained for New Hanover County and Miami-Dade County, respectively, from the ZTRAX database~\cite{zillow2018}.

\section{Roof-type classes}\label{roof class}

\noindent Gable and hip roofs are the most frequently used roof types for single-family dwellings in the United States~\cite{pita2008,model2005}. The roof shape of residential building stock was classified to be gable or hip, regardless of the variation of building shapes, in existing regional wind risk assessment frameworks (\eg HAZUS-MH~\cite{vickery2006} and FPHLM~\cite{pinelli2011}). Rectangular-shaped building archetypes with simple gable or hip roofs were adopted for the wind damage calculation. Nevertheless, most single-family residential buildings have non-rectangular footprints with resulting complex roof configurations. Wind tunnel tests~\cite{shao2018,parackal2016,uematsu2022} showed a substantial difference in the wind pressure magnitude and distribution on hip roofs with rectangular or non-rectangular building footprints. Moreover, Meng et al.~\cite{meng2023effects} assessed the wind vulnerability for roof sheathing on gable and hip roofs with different building shapes and observed significantly higher failure probabilities for non-rectangular roofs than those for rectangular roofs. Therefore, roof types used by existing wind vulnerability models may be oversimplified and can potentially cause underestimation in wind-induced damage and loss estimation.  

\begin{figure}[H]
\centering
\includegraphics[scale=0.58]{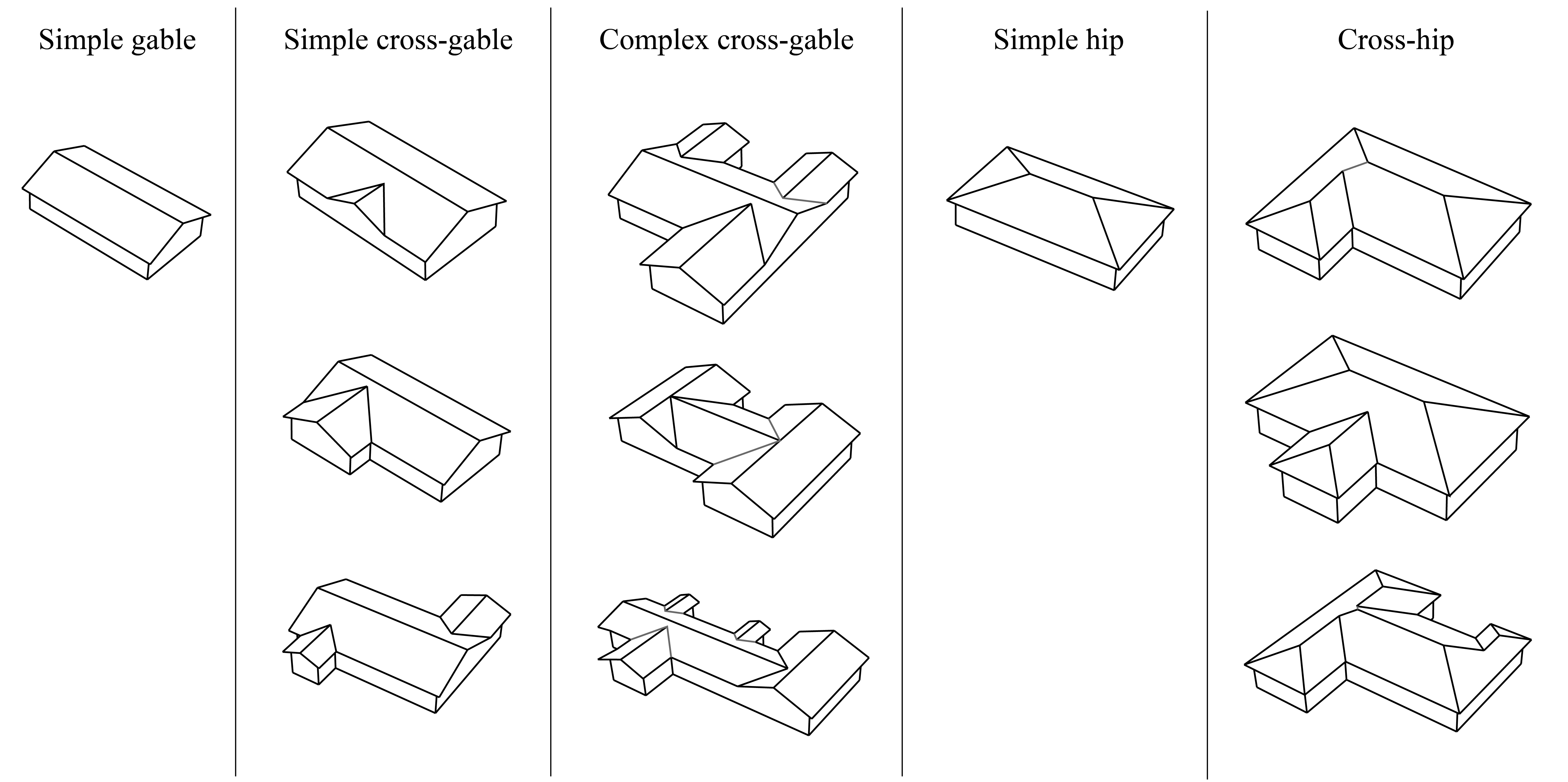}
\caption{Example building models with different types of roofs.}
\label{3d_model}
\end{figure}

Roof type classes developed by previous research~\cite{castagno2018,partovi2017,mohajeri2018,lingfors2017} were mainly for 3D city modeling and solar energy applications, not for the purpose of describing the wind resistance of the building. In existing roof class systems, many minor roof types (\eg mansard roof and shed roof) were considered, which could not be assessed in existing wind vulnerability models. Moreover, gable and hip roofs were usually not differentiated for complex roofs. To identify roofs with different wind resistance, a novel roof class system is proposed for the roof classification task based on the sensitivity analysis conducted by Meng et al.~\cite{meng2023effects}. Five roof types are defined, including simple gable, simple cross-gable, complex gable, simple hip, and cross-hip. The descriptions for the roof types are summarized in Table~\ref{table:rt}, and the example building models with each type of roof are depicted in Fig.~\ref{3d_model}. The proposed roof classes are intended to be representative of most single-family houses in the coastal area of the United States. Among the five roof types, gable and hip roofs with different roof complexity are differentiated, providing a more detailed and realistic description of the building stock compared to the roof types used by existing models. In addition, buildings with a combination of gable and hip roof sections are expected to be classified as one of the five roof types based on visual similarity.

\begin{table}[H]
\caption{Roof types for single-family houses.}
\label{table:rt}
\centering
\begin{tabular}{l c c}
\hline
\multicolumn{1}{l}{Roof type} &
\multicolumn{1}{c}{Notation} &
\multicolumn{1}{c}{Description}\\
\hline
Simple gable & g &
 Gable roof with one ridgeline\\
Simple cross-gable & scg &
 Gable roof with two or three ridgelines \\
 Complex cross-gable & ccg &
 Gable roof with more than three ridgelines \\
 Simple hip & h &
 Rectangular hip roof \\
 Cross-hip & ch &
 Non-rectangular hip roof \\
\hline
\end{tabular}
\end{table}

\section{Roof type classification using satellite imagery}\label{cnn}

\noindent A CNN model was created to classify roof types for single-family houses using individual-building-level satellite images. First, the construction of the training dataset and the process of training the CNN model are introduced in Section~\ref{train}. Next, the evaluation of the performance of the roof classification model on the testing datasets is shown in Section~\ref{test}. Finally, the roof-type data for the study areas created using the CNN model are presented in Section~\ref{data}.  

\subsection{Training}\label{train}
\begin{figure}[H]
\centering
\includegraphics[scale=0.38]{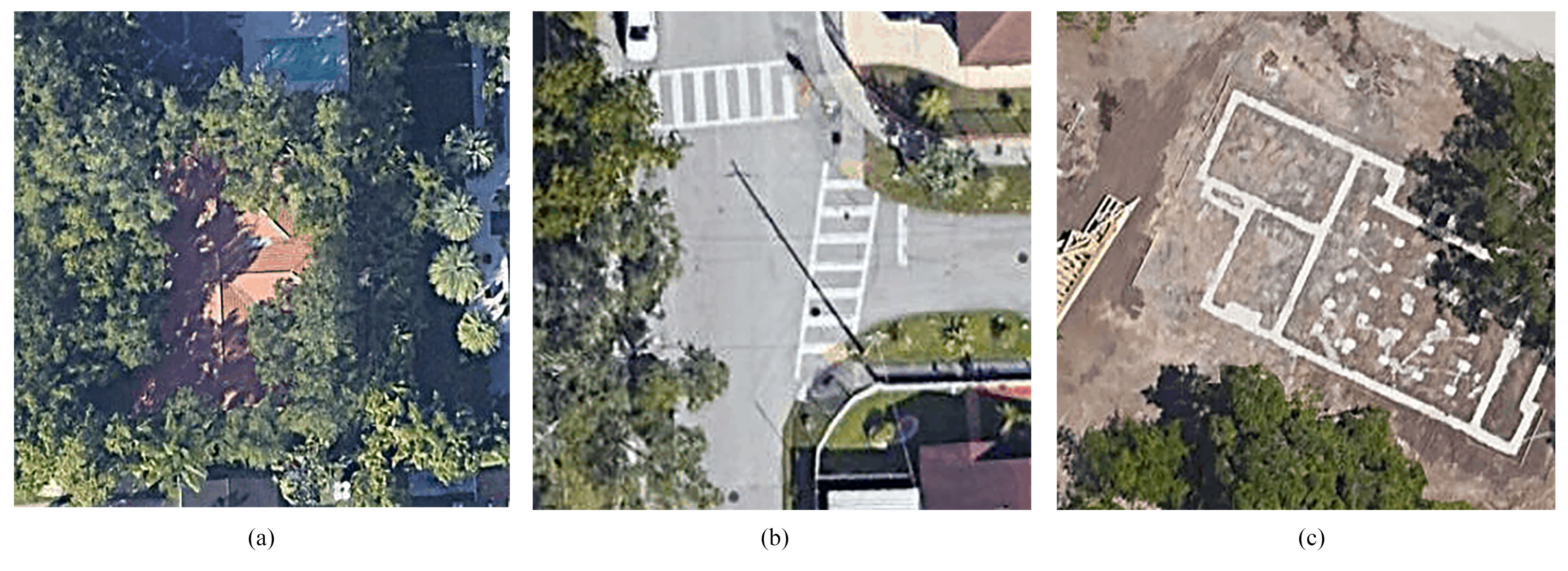}
\caption{Example low-quality satellite images: (a) Rooftop is heavily blocked by trees; (b) Geolocation of the building is incorrect; (c) The building was demolished.}
\label{nonroof}
\end{figure}

\begin{figure}[H]
\centering
\includegraphics[scale=0.25]{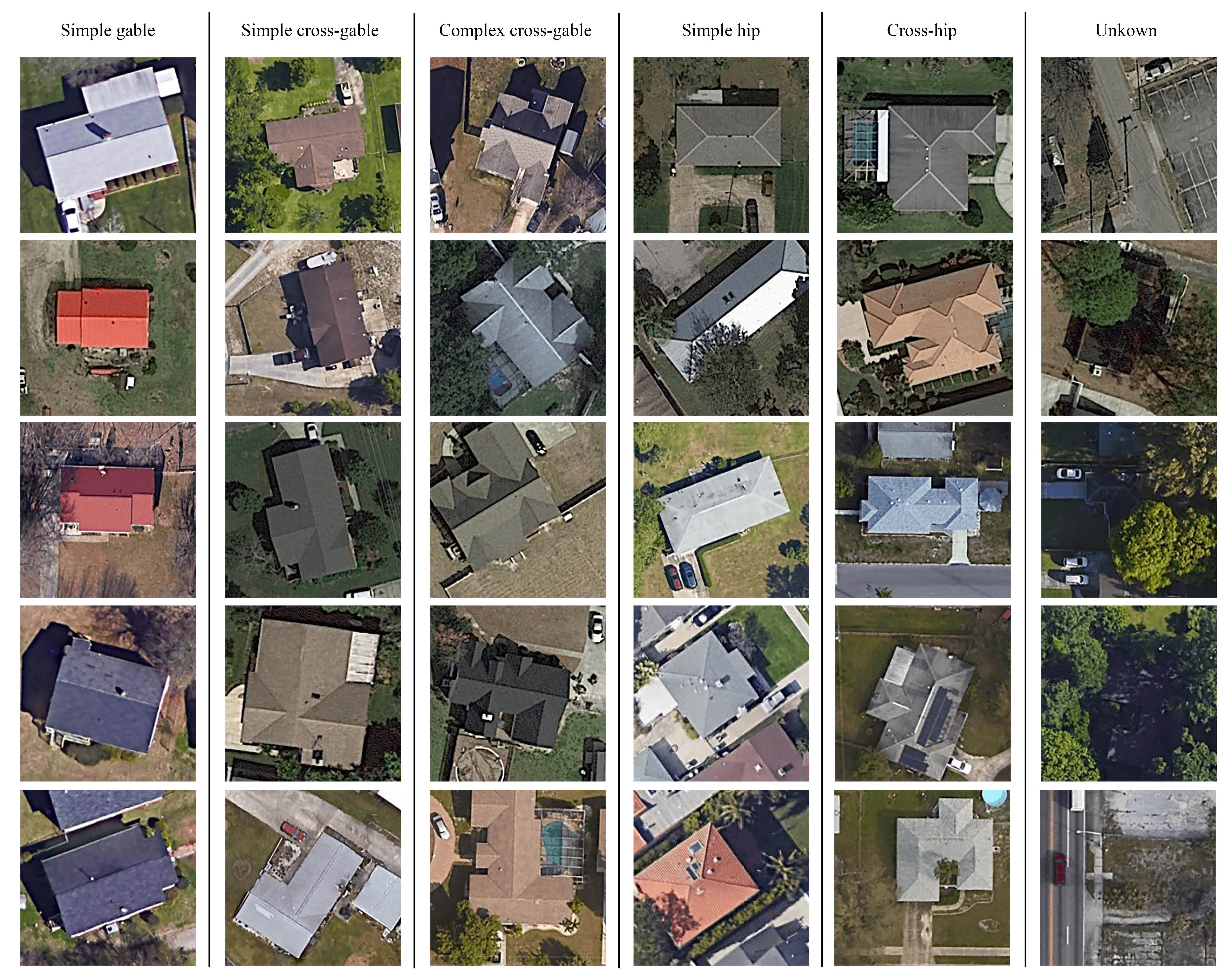}
\caption{Examples of training dataset for roof type classification.}
\label{training_image}
\end{figure}

\noindent A benchmark satellite image database was constructed to train the CNN model for the roof type classification. Following the image collection process described in Section~\ref{overview}, low-quality images that cannot provide sufficient information for rooftops may be obtained. As shown in Fig.~\ref{nonroof}, mainly three categories of images are considered to be low-quality: (1) unrecognizable roofs caused by tree occlusion, (2) mislocated images caused by incorrect building geolocation, and (3) demolished buildings. Among the three cases, tree occlusion is most frequently observed. To avoid erroneous roof data caused by low-quality images, an additional roof class, denoted as \textit{unknown}, is considered in the model. The roofs classified as unknown are predicted using data imputation algorithms described in Section~\ref{miss}. The dataset consists of 11,834 building-level satellite images collected from Google Maps for single-family houses on the eastern coast of the United States. The images were manually labeled and adjusted to ensure that each image only included one complete roof. Twenty percent of images for each category were randomly selected for the validation dataset. The number of training and validation images for each class are summarized in Table~\ref{table:rc image}. The representative satellite images for each roof class in the training dataset are shown in Fig.~\ref{training_image}.   

\begin{table}[H]
\caption{Distribution of training and validation images for the roof classification model.}
\label{table:rc image}
\centering
\begin{tabular}{l c c}
\hline
\multirow{2}{*}{Roof type}&
\multicolumn{2}{c}{Number of images}\\ \cline{2-3}
& Training & Validation\\
\hline
Simple gable & 1820 & 455\\
Simple cross-gable & 1805 & 451 \\
 Complex cross-gable & 1160 & 290 \\
 Simple hip & 1037 & 259 \\
 Cross-hip & 1823 & 456 \\
 Unknown & 1822 & 456 \\
\hline
\end{tabular}
\end{table}

\begin{figure}[H]
\centering
\includegraphics[scale=0.4]{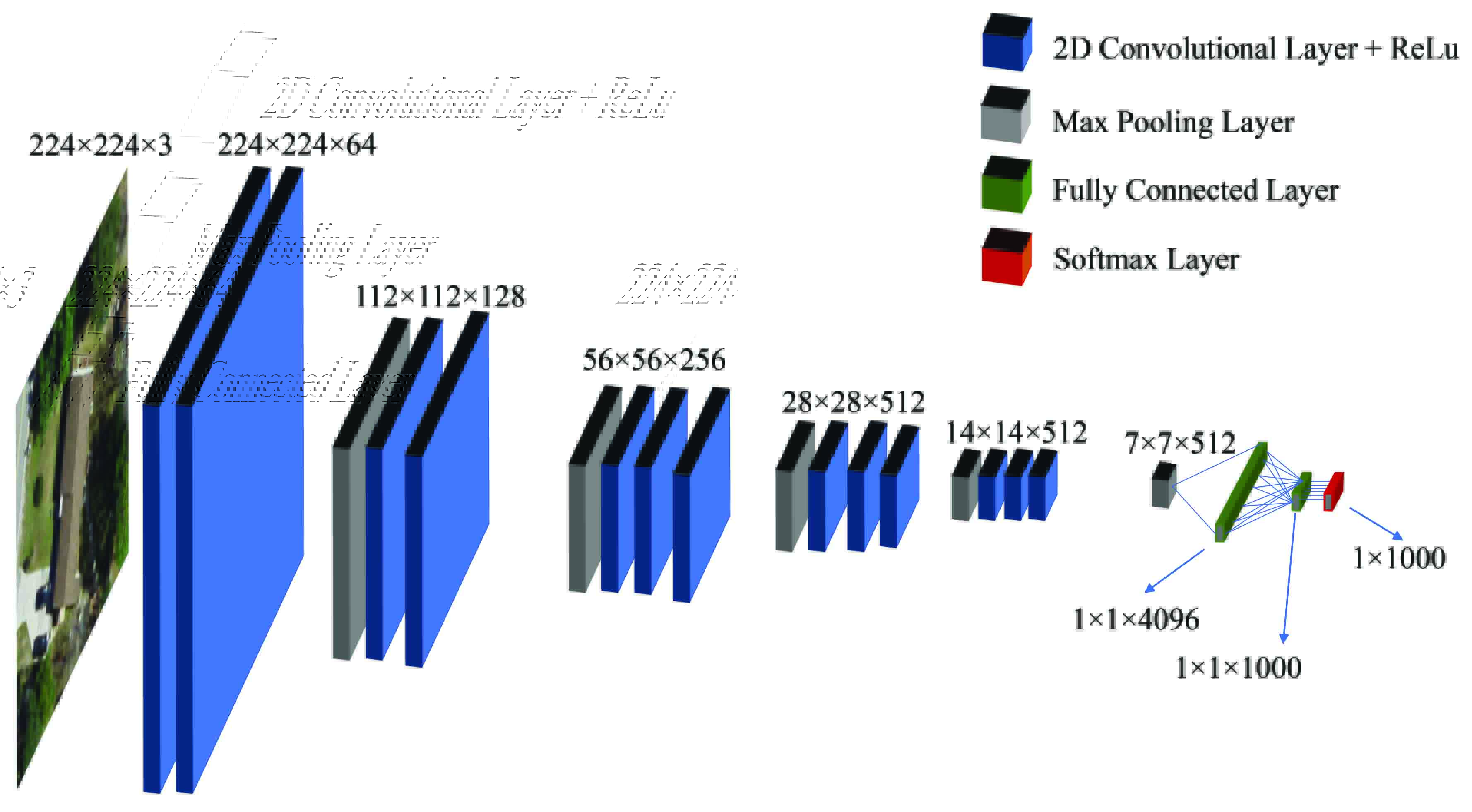}
\caption{VGG-19 architecture adopted from~\cite{simonyan2014}.}
\label{VGG}
\end{figure}

\begin{figure}[H]
\centering
\includegraphics[scale=0.22]{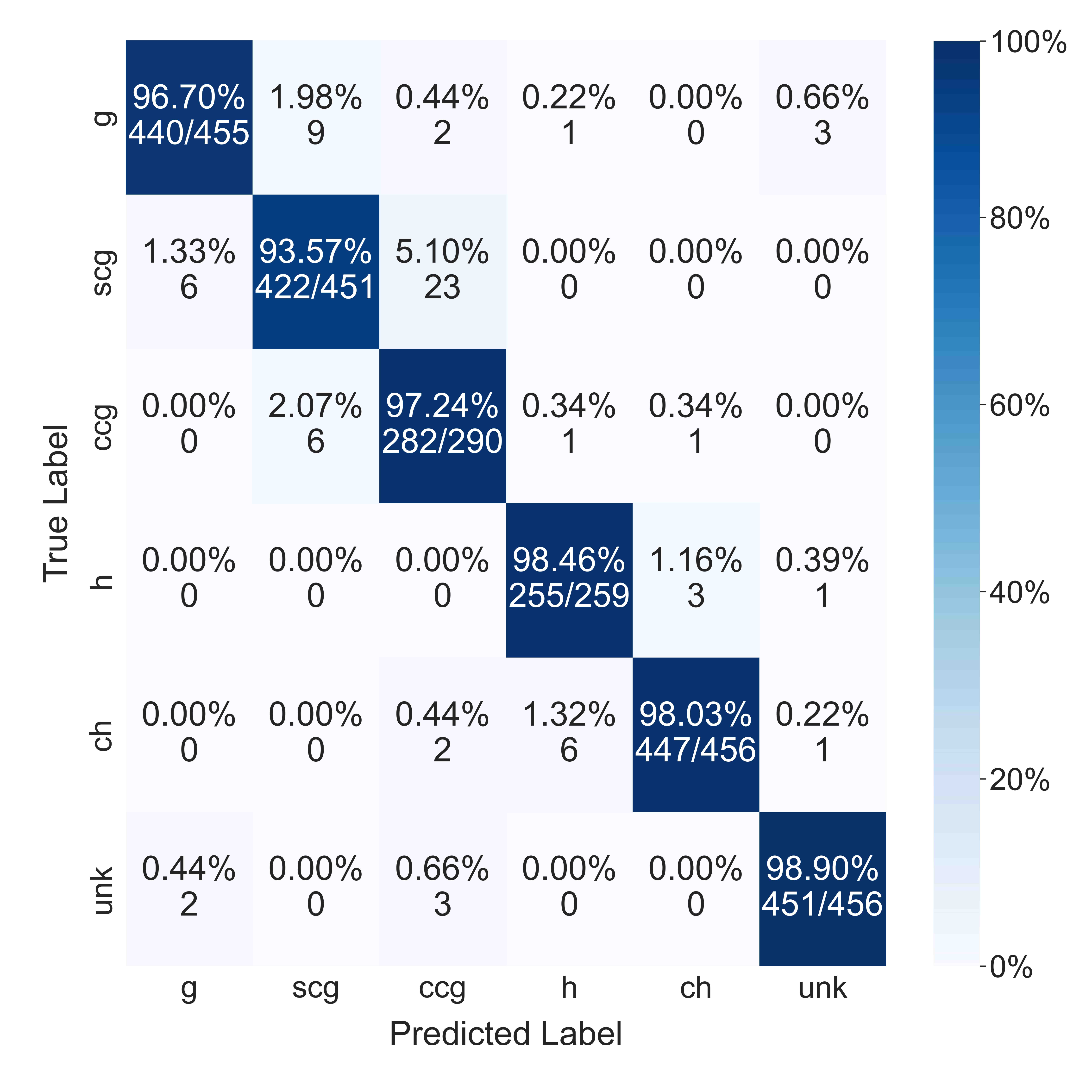}
\caption{Confusion matrix of VGG-19 evaluated on validation dataset.}
\label{confusion_valid}
\end{figure}

The generalizability (\ie accuracy regarding unseen test data) of neural networks is considered to be related to the representativeness of the training sets (\ie adequate samples to define the target data domain)~\cite{schwendicke2021,zhang2020}. Fine-tuning, a branch of Transfer Learning methods, is a viable tool to improve the generalizability of neural networks on image-related tasks, specifically with limited observations~\cite{BERGAMINI2020101052}. In this study, the roof classification model was obtained by fine-tuning a deep CNN architecture, VGG-19~\cite{simonyan2014} (Fig.~\ref{VGG}), that is pre-trained on the ImageNet dataset~\cite{Russakovsky2015} with more than 10 million images. The pre-trained model has a Softmax layer of size 1000, which is associated with the 1000 image categories in the ImageNet dataset. The VGG-19 architecture was proved to have the best performance on classification tasks on remote sensing data among the state-of-the-art CNN models, including roof classification using satellite images~\cite{buyukdemircioglu2021} and building use type classification using street view images~\cite{Kang2018}. Image augmentation techniques, including rotation, mirror, and adjustment of exposure and contrast, were applied to the training images to expand the training dataset and avoid overfitting. The images were re-scaled to the size of 224$\times$224 pixels per the VGGNet architecture. The batch size is set to 32 images. The stochastic gradient descent (SGD) optimizer is used for the optimization process, with the learning rate and momentum selected to be $1\times10^{-5}$ and 0.99, respectively. During training, the last fully connected layer and its Softmax layer (Fig.~\ref{VGG}) are modified to have 6 dimensions regarding the 6 roof type classes. First, the new fully connected layers and the output layer are trained on the training dataset for 40 epochs, and the rest of the network is used as a fixed feature extractor. After the training of the fully connected layers is complete, all layers of the network are fine-tuned for 16 epochs. Early stopping based on the validation accuracy is applied to the training process, and the model with highest validation accuracy is selected. The overall validation accuracy of the selected model is 97\%. The confusion matrix of the VGG-19 model evaluated on the validation dataset is shown in Fig.~\ref{confusion_valid}.

\subsection{Testing}\label{test}

\noindent Three testing datasets were constructed to provide a comprehensive evaluation of the model performance on buildings in different locations and with different roof styles. All testing images were manually labeled. The first two testing datasets---denoted as testing datasets A and B---consisted of buildings with pure gable, pure hip, and unknown roofs in two study areas. The third testing dataset---denoted as testing dataset C---is composed of buildings with a combination of gable and hip roof types while the dominant roof type is still differentiable.

\begin{figure}[H]
\centering
\includegraphics[scale=0.19]{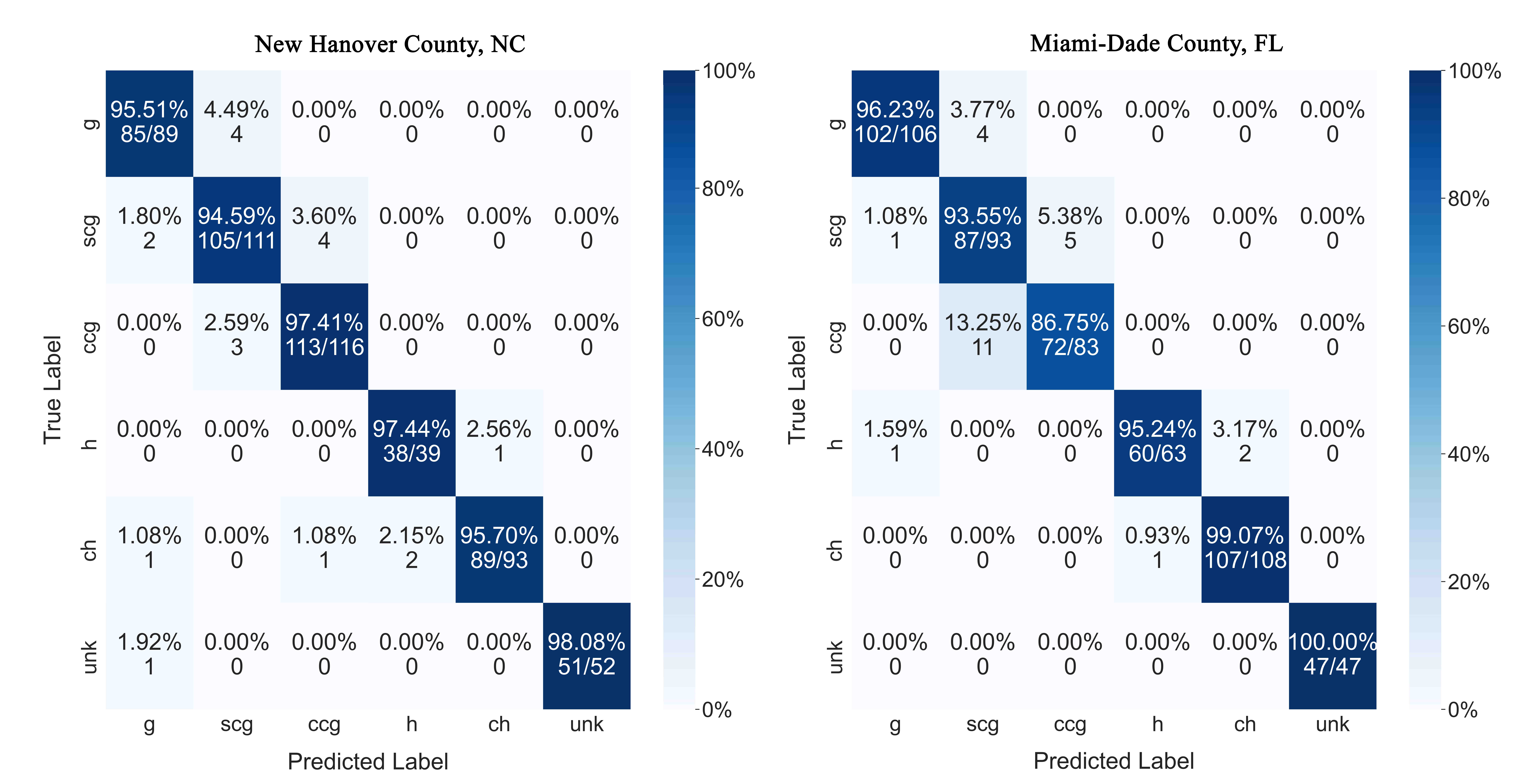}
\caption{Confusion matrix of VGG-19 evaluated on test datasets A and B (buildings with pure gable, pure hip, and unknown-type roofs).}
\label{roofconf}
\end{figure}

Testing datasets A and B, including 500 images each, were constructed using satellite images of single-family houses in the study areas, New Hanover County and Miami-Dade County. The confusion matrices of the VGG-19 model evaluated on testing datasets A and B are shown in Fig.~\ref{roofconf}. The precision, recall, and F1 score calculated for each class are summarized in Tables~\ref{table:score_NC} and~\ref{table:score_FL}. Overall F1 scores of 0.96 and 0.95 are achieved for testing buildings in New Hanover County and Miami-Dade County, respectively. As shown in Fig.~\ref{roofconf}, the model achieved high classification accuracy among all roof classes. Misclassification mainly happens between gable roofs with different complexity levels (\ie simplex gable, simple cross-gable, and complex cross-gable) that have high inter-class variance and intra-class similarity. Moreover, the model achieved the F1 scores of 0.99 and 1.00 when predicting images labeled unknown for the two study areas, proving the capability of the model to filter out low-quality images.  

\begin{table}[H]
\caption{Performance of VGG-19 on testing dataset A (buildings with pure gable, pure hip, and unknown-type roofs in New Hanover County, NC).}
\label{table:score_NC}
\centering
\begin{tabular*}{\textwidth}{l @{\extracolsep{\fill}} c c c c c}
\hline
Roof type & Precision & Recall & F1 score & Support \\
\hline
Simple gable & 
0.96 & 0.96 & 0.96 & 89\\
Simple cross-gable & 
0.94 & 0.95 & 0.94 & 111\\
Complex cross-gable & 
0.96 & 0.97 & 0.97 & 116\\
Simple hip & 
0.95 & 0.97 & 0.96 & 39\\
Cross-hip & 
0.99 & 0.96 & 0.97 & 93\\
Unknown &
1.00 & 0.98 & 0.99 & 52\\
Overall &
0.96 & 0.96 & 0.96 & 500\\
\hline
\end{tabular*}
\end{table}

\begin{table}[H]
\caption{Performance of VGG-19 on testing dataset B (buildings with pure gable, pure hip, and unknown-type roofs in Miami-Dade County, FL).}
\label{table:score_FL}
\centering
\begin{tabular*}{\textwidth}{l @{\extracolsep{\fill}} c c c c c}
\hline
Roof type & Precision & Recall & F1 score & Support \\
\hline
Simple gable & 
0.98 & 0.96 & 0.97 & 106\\
Simple cross-gable & 
0.85 & 0.94 & 0.89 & 93\\
Complex cross-gable & 
0.94 & 0.87 & 0.90 & 83\\
Simple hip & 
0.98 & 0.95 & 0.97 & 63\\
Cross-hip & 
0.98 & 0.99 & 0.99 & 108\\
Unknown &
1.00 & 1.00 & 1.00 & 47\\
Overall &
0.95 & 0.95 & 0.95 & 500\\
\hline
\end{tabular*}
\end{table}

\begin{figure}[H]
\centering
\includegraphics[scale=0.3]{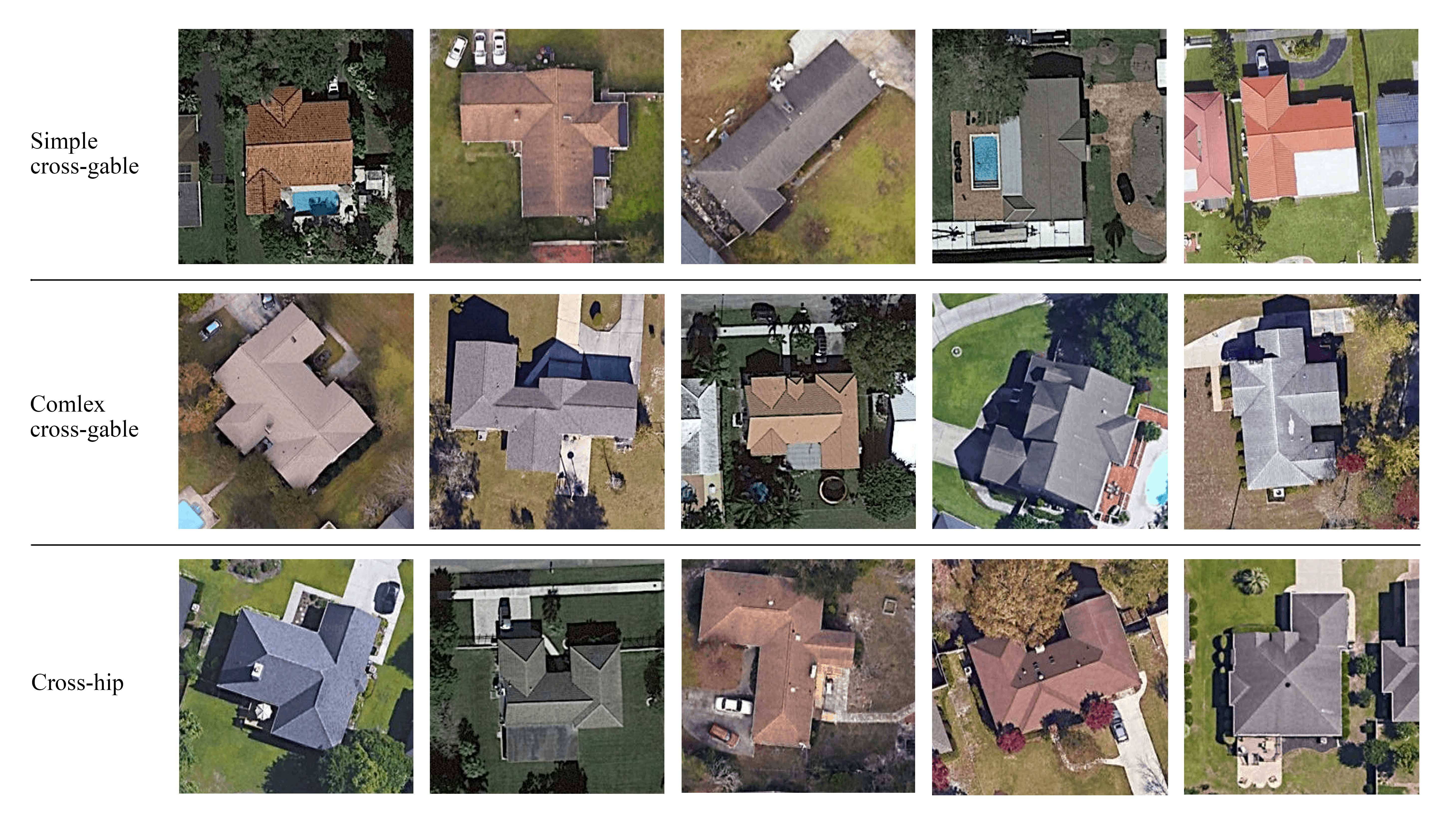}
\caption{Examples of testing dataset C (buildings with complex and mixed roof shapes).}
\label{test_image_mix}
\end{figure}

\begin{figure}[H]
\centering
\includegraphics[scale=0.2]{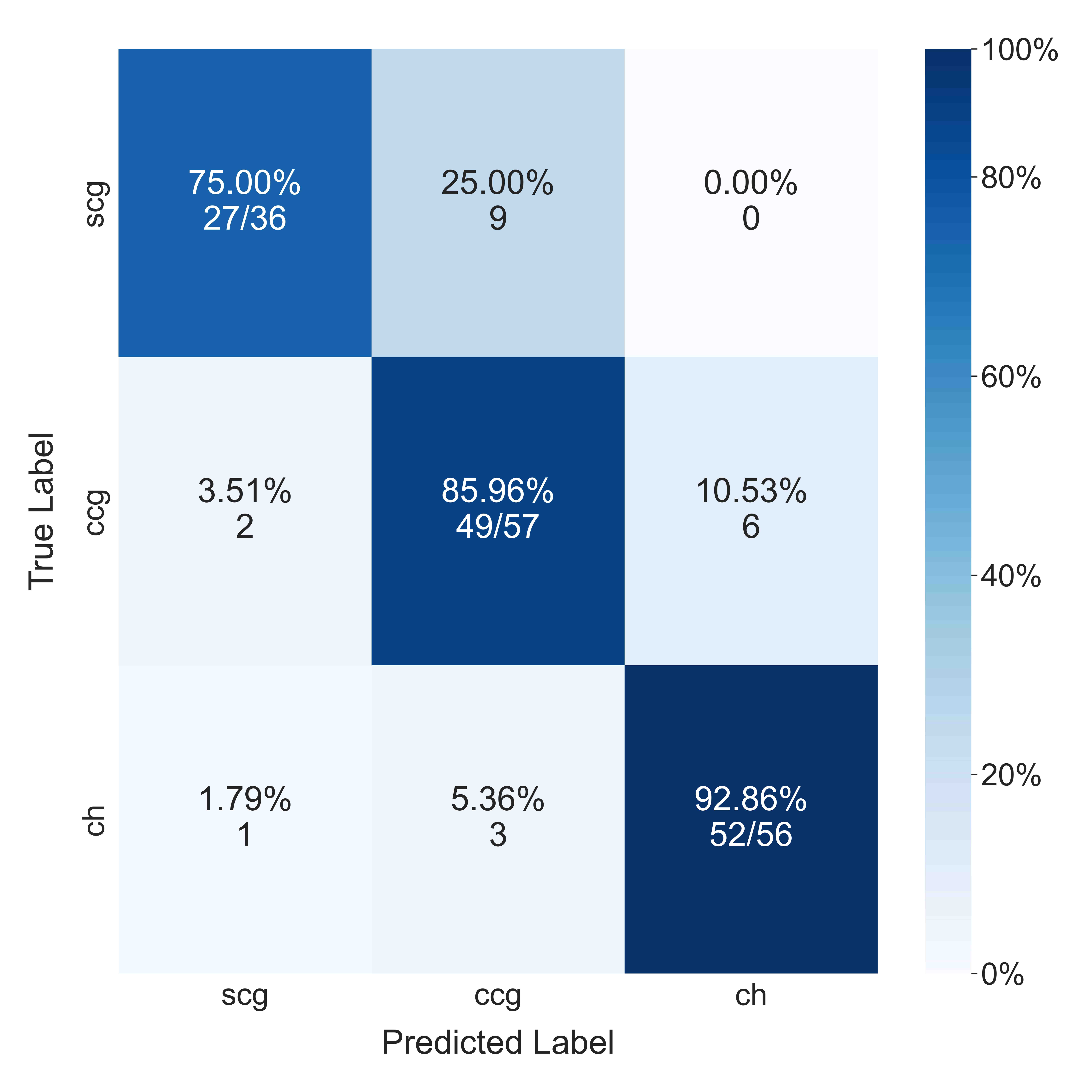}
\caption{Confusion matrix of VGG-19 evaluated on testing dataset C (buildings with mixed roof shapes).}
\label{confusion_mix}
\end{figure}

Single-family houses with non-rectangular building shapes are often observed to have a combination of gable and hip roof shapes. However, mix-shaped roofs cannot be modeled using the existing wind risk assessment methodology. Following the roof type classification principles suggested by HAZUS-MH~\cite{vickery2006}, buildings with mixed roof shapes should be classified using the dominant roof type available in building archetypes. To assess the performance of the CNN model on buildings with mixed roof shapes, testing dataset C, including 150 images for mix-shaped roofs, was created for houses in New Hanover County and Miami-Dade County. Rectangular-shaped roofs with mixed roof shapes are rare in practice, and the dominant roof type for such roofs is hard to differentiate. Therefore, only non-rectangular houses were considered in the testing dataset. Each image was labeled with one of the complex roof classes (\ie simple cross-gable, complex cross-gable, and cross-hip) based on the dominant roof shape. Example images for each roof class are shown in Fig.~\ref{test_image_mix}. The confusion matrix of the CNN model evaluated on the testing dataset C is depicted in Fig.~\ref{confusion_mix}, and the precision, recall, and F1 score for each class are listed in Table~\ref{table:score_mix}. As shown in Fig.~\ref{confusion_mix}, the prediction accuracy for mix-shaped roofs is lower than roofs with pure gable or hip roof styles. But the dominant roof type is still correctly identified for most mix-shaped roofs. In summary, the testing results prove the capability of the CNN model to generate reliable and reasonable roof-type data using satellite images.      

\begin{table}[H]
\caption{Performance of VGG-19 on testing dataset C (buildings with mixed roof shapes).}
\label{table:score_mix}
\centering
\begin{tabular*}{\textwidth}{l @{\extracolsep{\fill}} c c c c c}
\hline
Roof type & Precision & Recall & F1 score & Support \\
\hline
Simple cross-gable & 
0.90 & 0.75 & 0.82 & 36\\
Complex cross-gable & 
0.80 & 0.86 & 0.83 & 57\\
Cross-hip & 
0.90 & 0.91 & 0.90 & 57\\
Overall &
0.86 & 0.85 & 0.86 & 150\\
\hline
\end{tabular*}
\end{table}

\subsection{Roof-type dataset for study areas}\label{data}

\begin{figure}[H]
\centering
\includegraphics[scale=0.3]{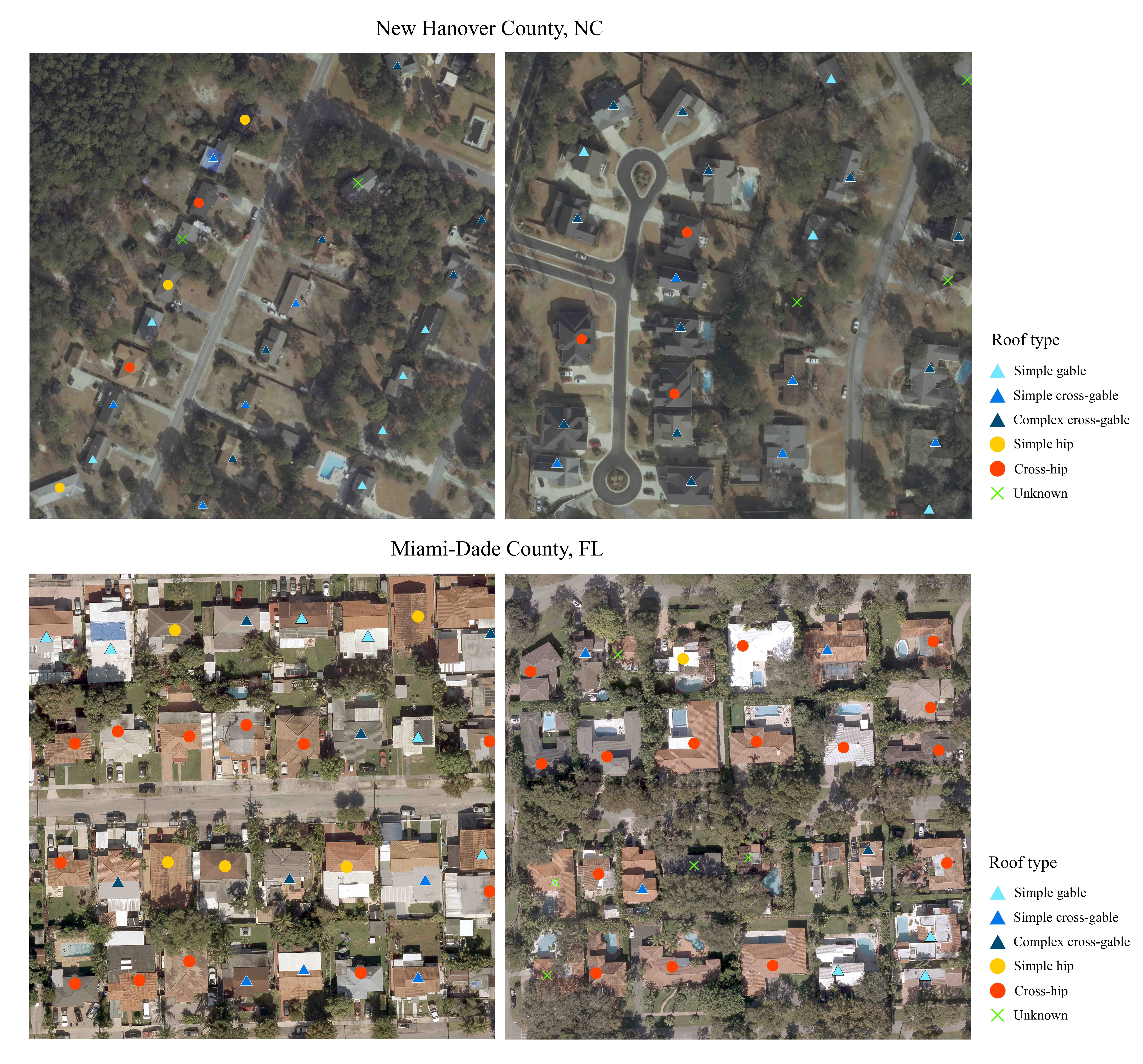}
\caption{Predicted roof type map for single-family houses in study areas. Roof-type labels are shown at building centroids obtained using Google Geocoding API. The neighborhood-level satellite images are for illustrative purposes only, not the images used for roof type classification.}
\label{sat_map}
\end{figure}

\begin{figure}[H]
\centering
\includegraphics[scale=0.5]{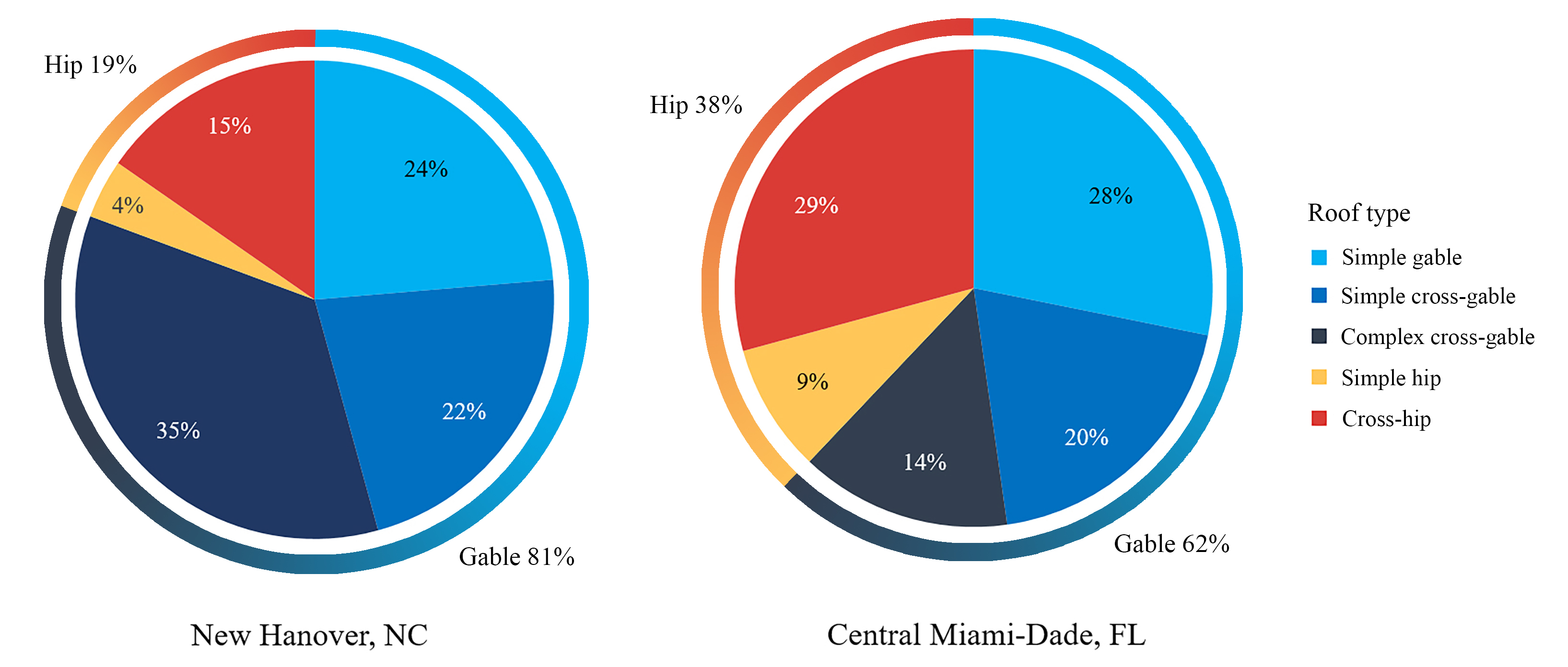}
\caption{Roof type distribution for study areas.}
\label{roof_dist}
\end{figure}

\noindent The trained CNN model was applied to the study areas to generate large-scale roof-type dataset for single-family houses. Following the image collection procedure described in Section~\ref{overview}, a satellite image for each building was downloaded from Google Maps and classified using the CNN model.  The predicted neighborhood-level roof type map is shown in Fig.~\ref{sat_map}. Valid roof-type data (\ie roofs not classified as unknown) was obtained for more than 80\% of buildings. Roof types for 17\% and 11\% of houses in New Hanover County and Miami-Dade County were classified as unknown, and the city-level roof type distribution for the rest of the houses is summarized in Fig.~\ref{roof_dist}. It can be seen in Fig.~\ref{roof_dist} that gable roof is the dominant roof type in both areas, which is consistent with the roof-type data collected by previous studies~\cite{vickery2006,Crandell1993}. The proportion of houses with hip roofs is 19\% higher for New Hanover County than those for Miami-Dade County. The ratio between gable and hip roof in Miami-Dade County is 1.6:1, which is close to the ratio of 2:1 derived from tax appraisers' databases for Brevard County and Escambia County in Florida~\cite{model2005}. Regarding roof complexity, 72\% and 63\% of houses have complex roofs (\ie simple cross-gable, complex cross-gable, and cross-hip) in New Hanover County and Miami-Dade County, respectively. On the other hand, a considerable amount of houses has simple rectangular roofs, and simple gable roof is the second most frequently observed roof type in both study areas.  

\begin{figure}[H]
\centering
\includegraphics[scale=0.32]{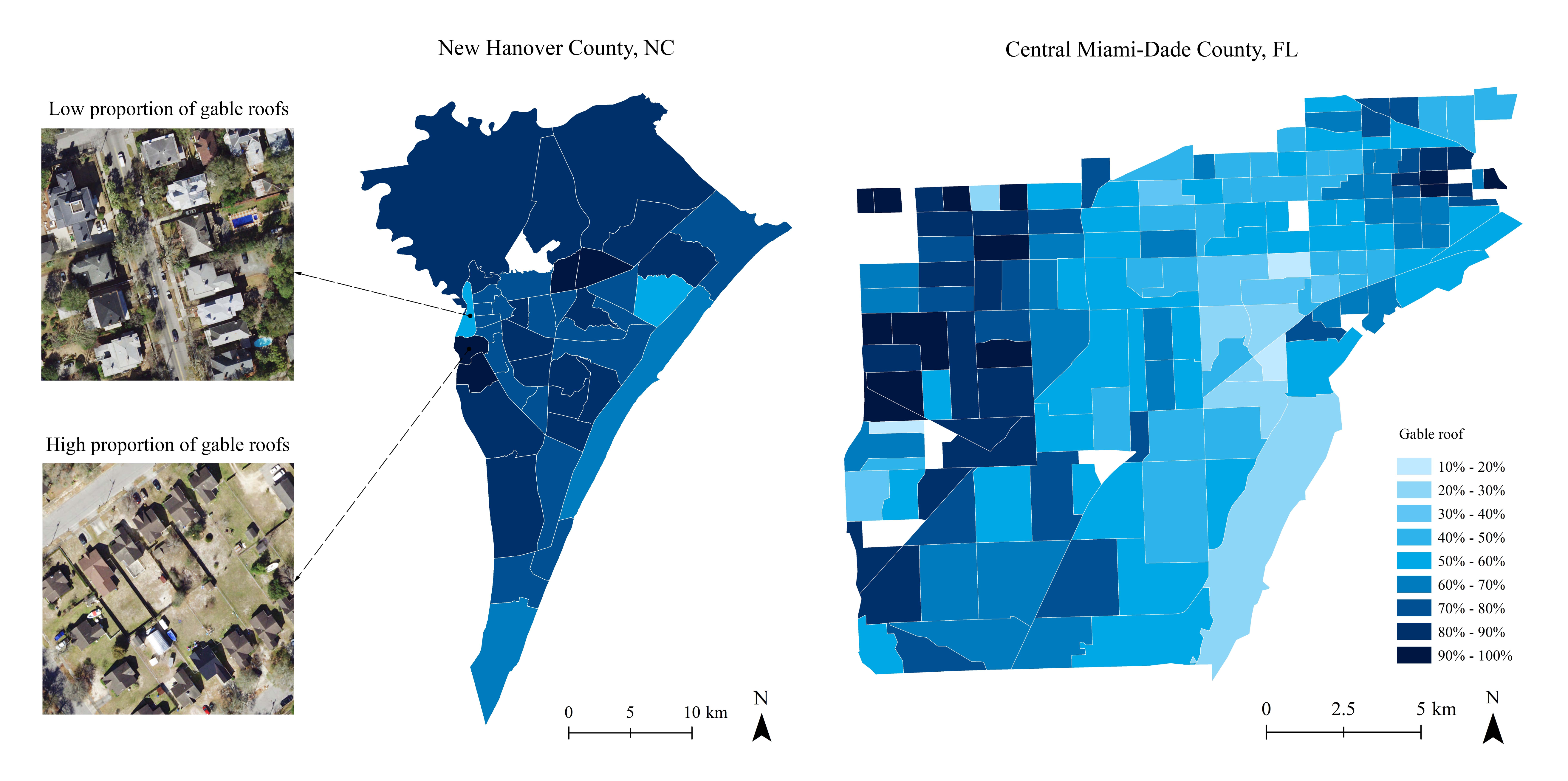}
\caption{Proportion of single-family houses with gable roofs per census tract in study areas. (census tracts with less than ten houses were removed from the map)}
\label{gable_ratio}
\end{figure}

\begin{figure}[H]
\centering
\includegraphics[scale=0.31]{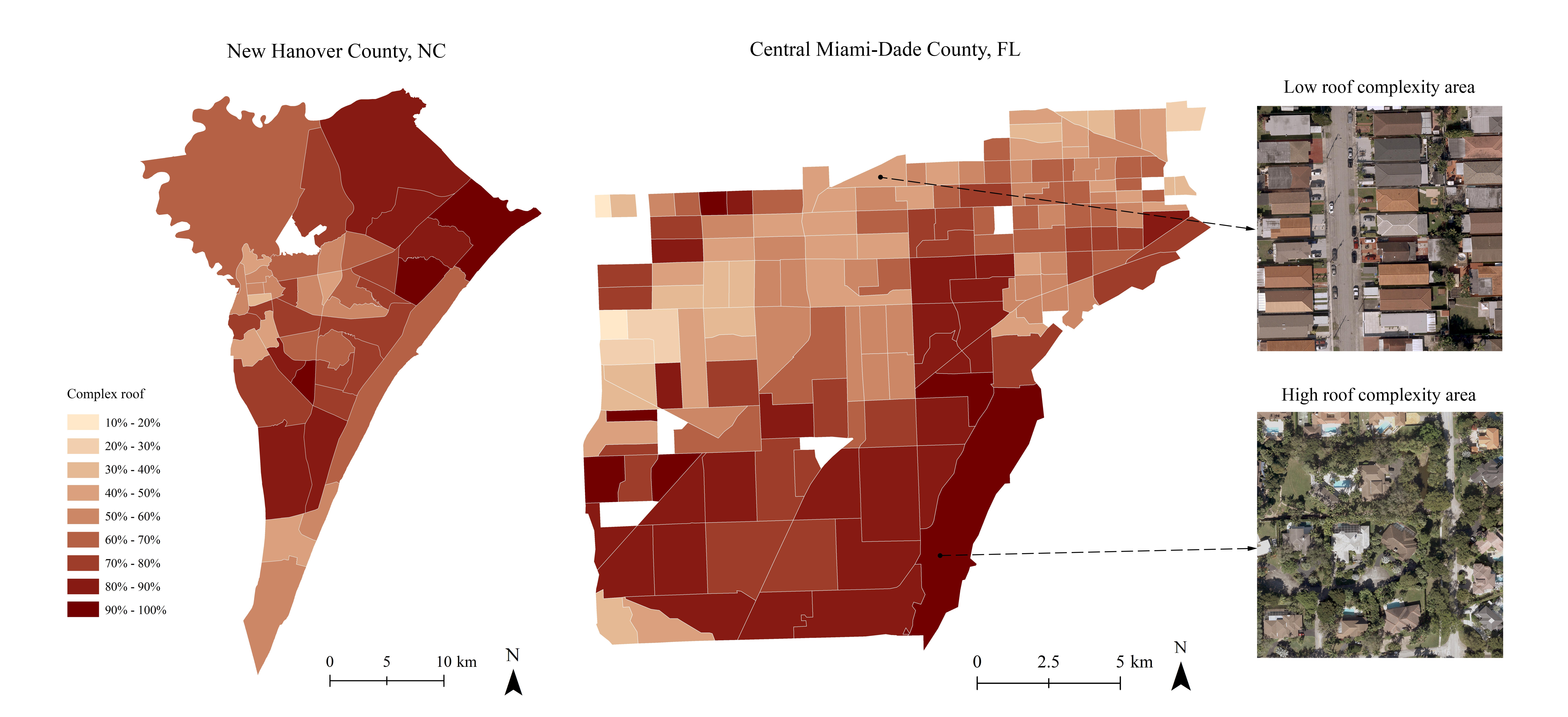}
\caption{Proportion of single-family houses with complex roofs, including simple cross-gable, complex cross-gable and cross-hip roofs, per census tract in study areas. (census tracts with less than ten houses were removed from the map)}
\label{comp_ratio}
\end{figure}

A census tract is often chosen as the geographic area unit for regional wind risk assessments. The distribution of building attributes, including roof type, is usually determined for each census tract to model the building stock. To investigate the distribution of roof type in the level of census tracts, the proportions of single-family houses with gable roofs (\ie simple gable, simple cross-gable, and complex cross-gable) and complex roofs (\ie simple cross-gable, complex cross-gable, and cross-hip) were calculated for each census tract in study areas and plotted in Figs.~\ref{gable_ratio} and~\ref{comp_ratio}. As shown in Fig.~\ref{gable_ratio}, gable roof is the dominant roof type in all census tracts in New Hanover County, with the proportion of houses with gable roofs varying from 55\% to 92\%. On the contrary, in central Miami-Dade County, hip roof is the predominant roof type for 26\% of census tracts. More than 80\% of single-family houses have hip roofs in three census tracts. Regarding roof complexity, the proportion of houses with complex roofs varies between 39\% and 97\% in New Hanover County and between 16\% and 97\% in Miami-Dade County. Significant spatial variance in roof type distribution is shown at both city and census tract levels. Adjacent census tracts can have completely different roof type distribution. Therefore, detailed roof-type data are crucial for the accuracy of wind vulnerability assessments.    

\section{Missing roof data population}\label{miss}

\begin{figure}[H]
\centering
\includegraphics[scale=0.5]{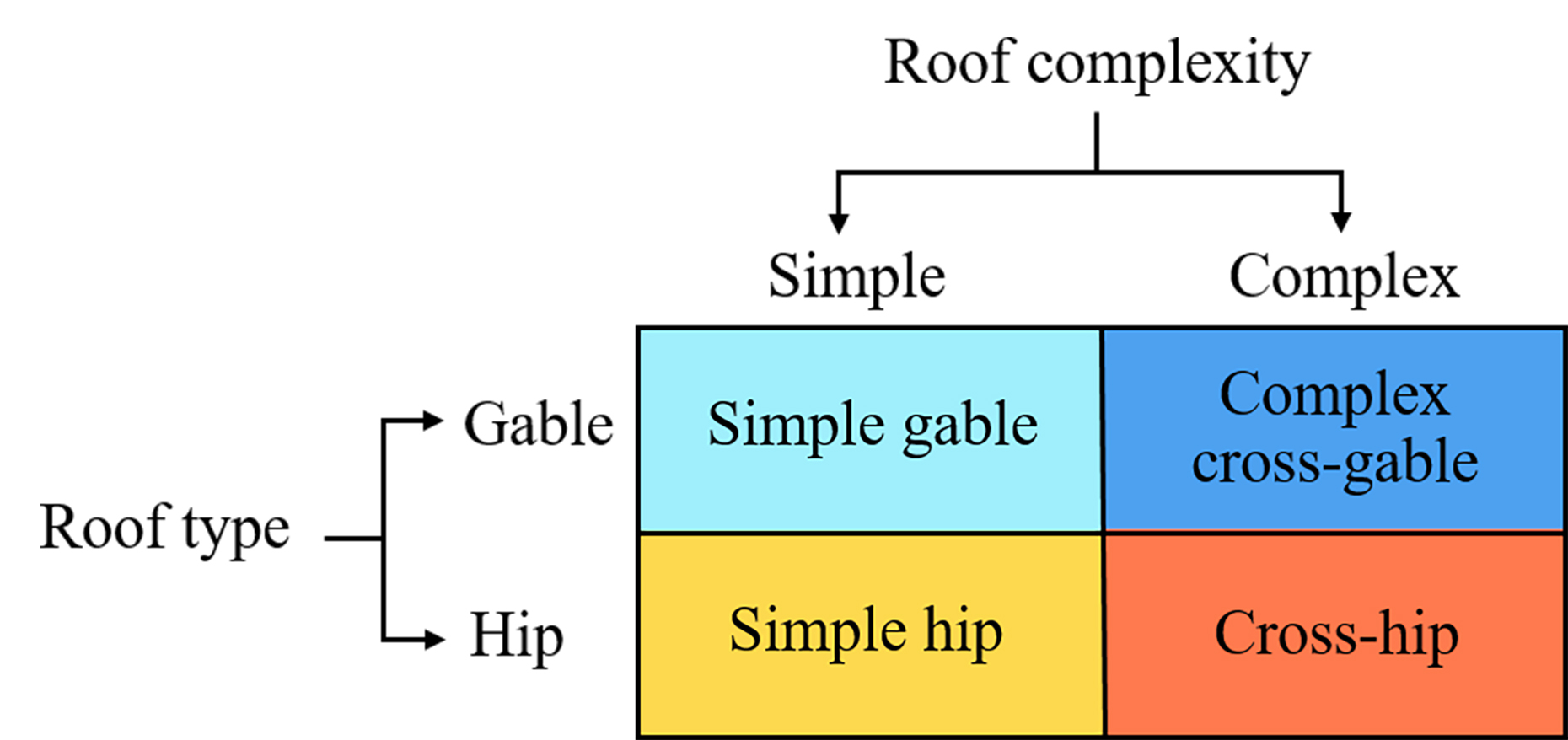}
\caption{Process of mapping roof features predicted by imputation algorithms to roof types. For simplification, complex cross-gable roofs are used for representing all complex gable roofs (\ie simple cross-gable and complex cross-gable).}
\label{mapping}
\end{figure}

\noindent As discussed in Section~\ref{cnn}, low-quality images are identified by the CNN model, resulting in missing roof-type data in the building inventory. Roof type was found to be correlated with other building characteristics (\eg year built and building area)~\cite{pita2011,Hamid2021}. To improve the completeness of the database and support more detailed analysis (\eg neighborhood-level analysis), data imputation algorithms were developed to populate missing roof-type data. Valid roof-type data were obtained for 56,863 and 76,045 single-family houses in New Hanover County and Miami-Dade County using the CNN model, which is considered as the ground truth. Other critical building characteristics, including year built, building value, building area, and number of stories, were obtained from the ZTRAX database~\cite{zillow2018}. Random Forest and Support Vector Machine were used for imputing missing data, which are frequently used machine learning models for building-related classification tasks~\cite{mohajeri2018,castagno2018,taghinezhad2020,soares2021}. Based on wind tunnel tests~\cite{shao2018, parackal2016}, roof type to be gable or hip has a more significant effect on wind pressures acting on the roof than the building shape. Therefore, different models were trained to predict roof type (gable/hip) and roof complexity (simple/complex) to simplify the classification task to binary classification and avoid cumulative errors. Figure~\ref{mapping} shows how roof features predicted by the imputation models are mapped to the roof types defined in Table~\ref{table:rt}.

\begin{figure}[H]
\centering
\includegraphics[scale=0.35]{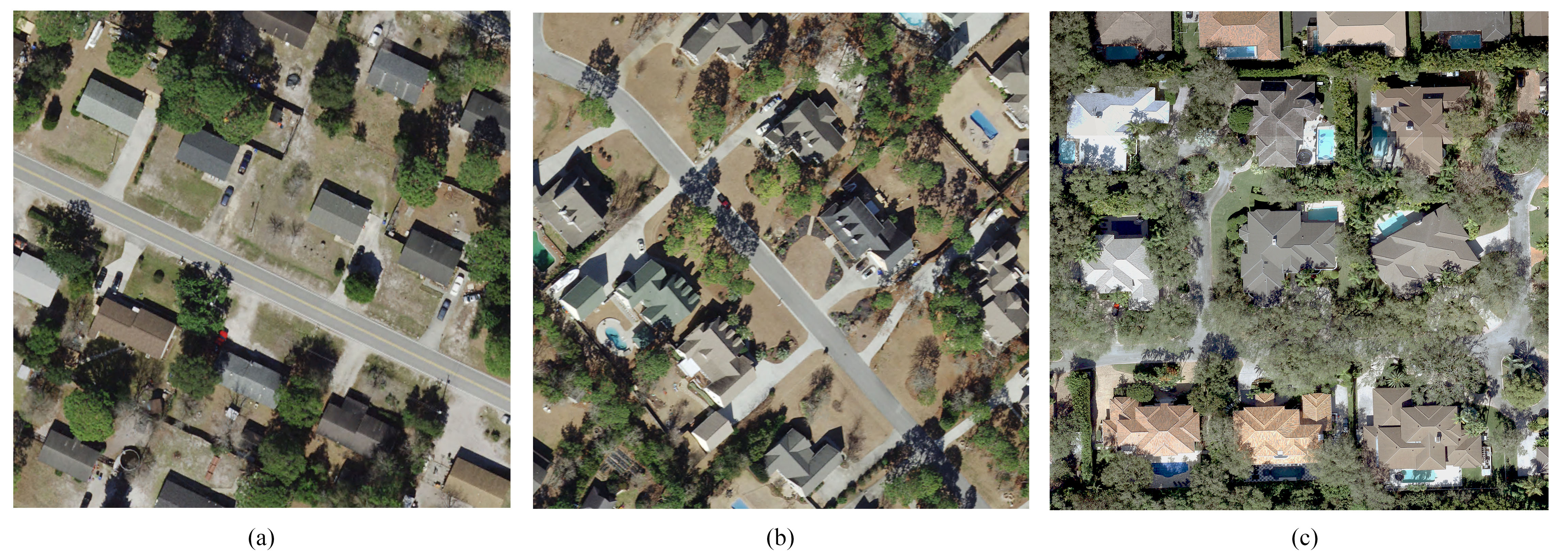}
\caption{Neighborhoods with similar roof styles among the single-family houses: (a) neighborhood with all simple gable roofs; (b) neighborhood with all complex cross-gable roofs; (c) neighborhood with all cross-hip roofs.}
\label{similar_neighbor}
\end{figure}

\begin{figure}[H]
\centering
\includegraphics[scale=0.5]{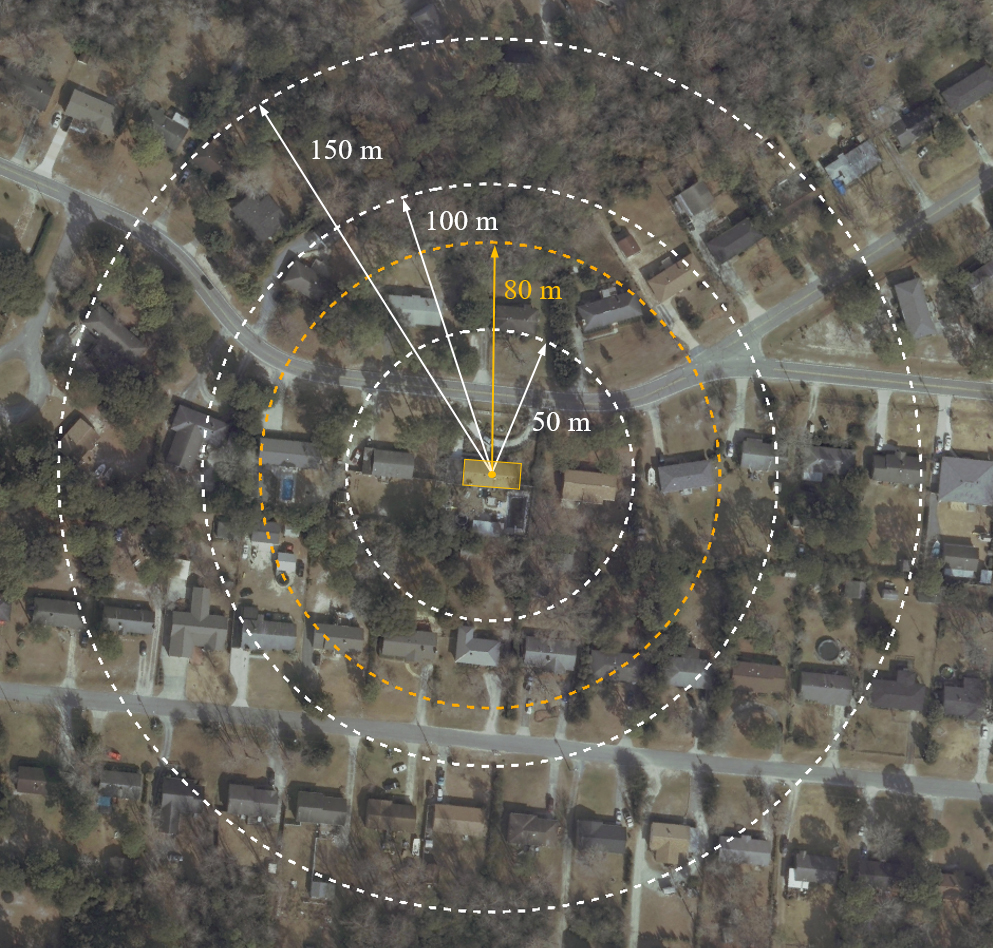}
\caption{Candidate search radius used to calculate neighborhood-level roof type distribution for predicting missing roof-type data.}
\label{buffer}
\end{figure}

Besides critical building characteristics, the spatial correlation of roof types is also considered to improve the performance of imputation models. As shown in Fig.~\ref{similar_neighbor}, single-family houses in the same neighborhood tend to have similar roof styles. Therefore, roof type of a given building can be estimated based on the roof types of its neighboring buildings. Four values of search radius, 50m, 80m, 100m, and 150m (Fig.~\ref{buffer}), were tested to find the optimal search radius for the neighboring buildings. The roof type (gable/hip) for each house in the building database is predicted using the dominant roof type (gable/hip) of the neighboring houses located within the search radius. For example, if more than 50\% of neighbors for a given building have hip roofs, that building is predicted to have a hip roof. Prediction accuracy was calculated based on the houses with neighboring houses available, and the proportion of homes without any neighbor within the search radius was counted. It can be seen in Table~\ref{table:select buffer} that the prediction accuracy decreases with a larger search radius, while a smaller search radius results in a higher likelihood of failing to find neighbors. Consequently, a search radius of 80m is selected to ensure high prediction accuracy and data availability. The \textit{neighbor type} and the \textit{neighbor complexity}, which represents the proportion of neighbors with gable roofs (\ie simple gable, simple cross-gable, and complex cross-gable) and with complex roofs (\ie simple cross-gable, complex cross-gable, and cross-hip), are calculated for each building and used as input features for training the machine learning models. The missing neighboring roof features were imputed using the mean values in the corresponding study area.

\begin{table}[H]
\caption{Prediction accuracy of roof type (gable or hip) using dominant roof type surrounding each building, and the proportion of buildings missing neighbors within the search radius.}
\label{table:select buffer}
\centering
\begin{tabular}{l c c c c}
\hline
\multirow{2}{*}{Search radius (m)}&
\multicolumn{2}{c}{New Hanover}  &
\multicolumn{2}{c}{Miami-Dade}\\ \cline{2-5}
& Accuracy & Missing data & Accuracy & Missing data\\
\hline
50 & 86.8\% & 6.5\% & 77.4\% & 4.3\%\\
\textbf{80} & \textbf{84.9\%} & \textbf{1.6\%} & \textbf{76.1\%} & \textbf{0.7\%}\\
100 & 84.4\% & 0.9\% & 74.6\% & 0.2\%\\
150 & 83.4\% & 0.5\% & 72.7\% & 0.2\%\\
\hline
\end{tabular}
\end{table}

\begin{figure}[H]
\centering
\includegraphics[scale=0.65]{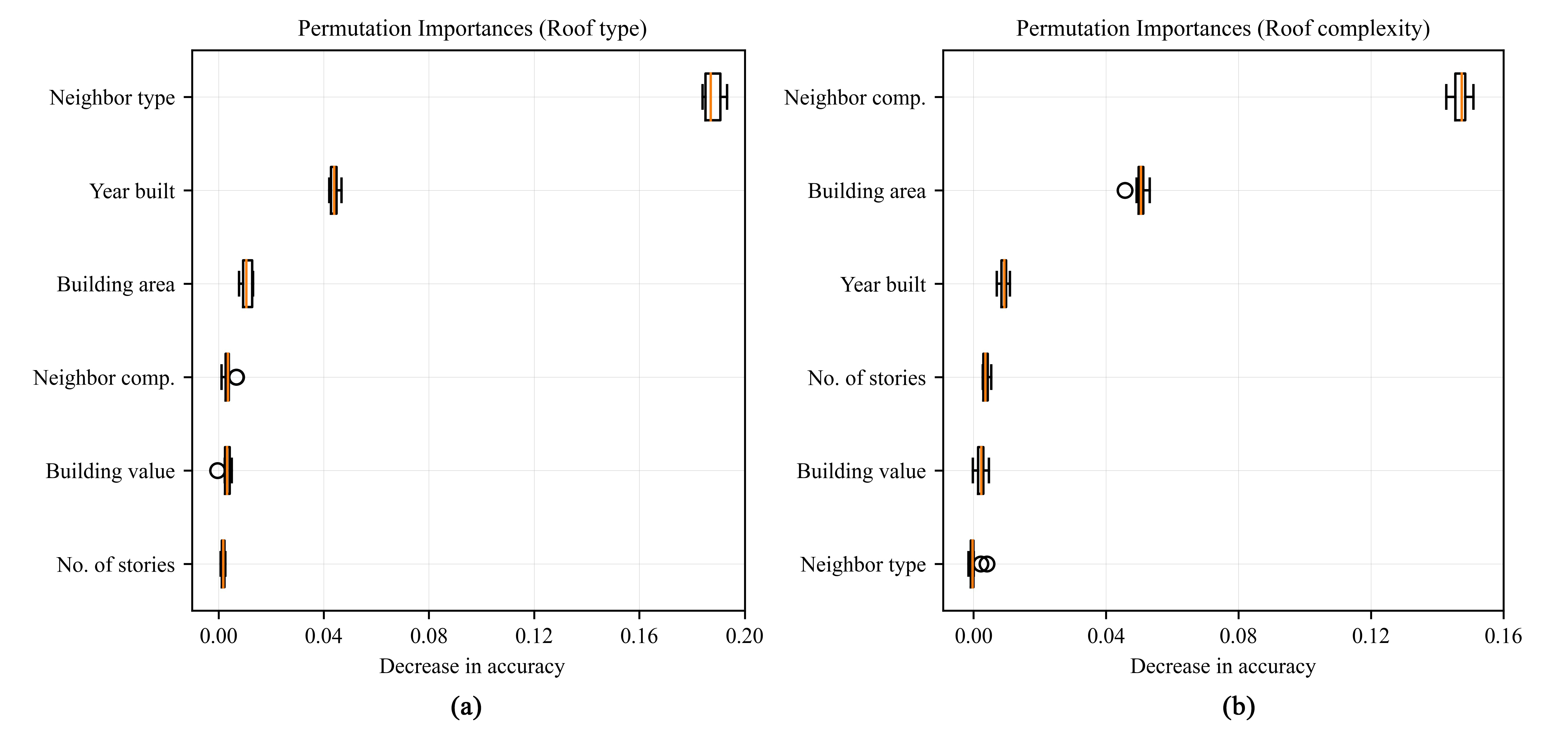}
\caption{Permutation feature importance evaluated on validation dataset for Miami-Dade County with Random Forest : (a) Predicting roof to be gable or hip; (b) Predicting roof to be simple or complex.}
\label{importance}
\end{figure}

To select the most relevant features for predicting roof type, permutation feature importance was evaluated for all available building attributes, which measures the decrease in prediction accuracy when permuting a feature. Building database for each study area is split into 75\% for training and 25\% for validation. Permutation importance for each feature was calculated on the validation dataset for each model. Plots of permutation importance calculated based on Random Forest for buildings in Miami-Dade County are shown in Fig.~\ref{importance} as an example. As expected, the roof style of neighboring buildings plays the most important role in roof type prediction. The \emph{year built} and the \emph{building area} are also found to be relevant for predicting the roof type, which is in agreement with conclusions drawn by Pinelli et al.~\cite{Hamid2021}. The features selected for predicting roof type and roof complexity are summarized in Table~\ref{table:feature}.

\begin{table}[H]
\caption{Features used for roof type and roof complexity prediction with data imputation algorithms.}
\label{table:feature}
\centering
\small
\renewcommand{\arraystretch}{1.25}
\begin{tabular}{l l l}
\hline
\multicolumn{1}{l}{Property} &
\multicolumn{1}{l}{Category} &
\multicolumn{1}{l}{Feature} \\
\hline
Roof type & Gable/hip & Year built, building area, neighbor roof type \\
Roof complexity & Simple/complex & Year built, building area, neighbor roof complexity \\
\hline
\end{tabular}
\normalsize
\end{table}

\begin{figure}[H]
\centering
\includegraphics[scale=0.4]{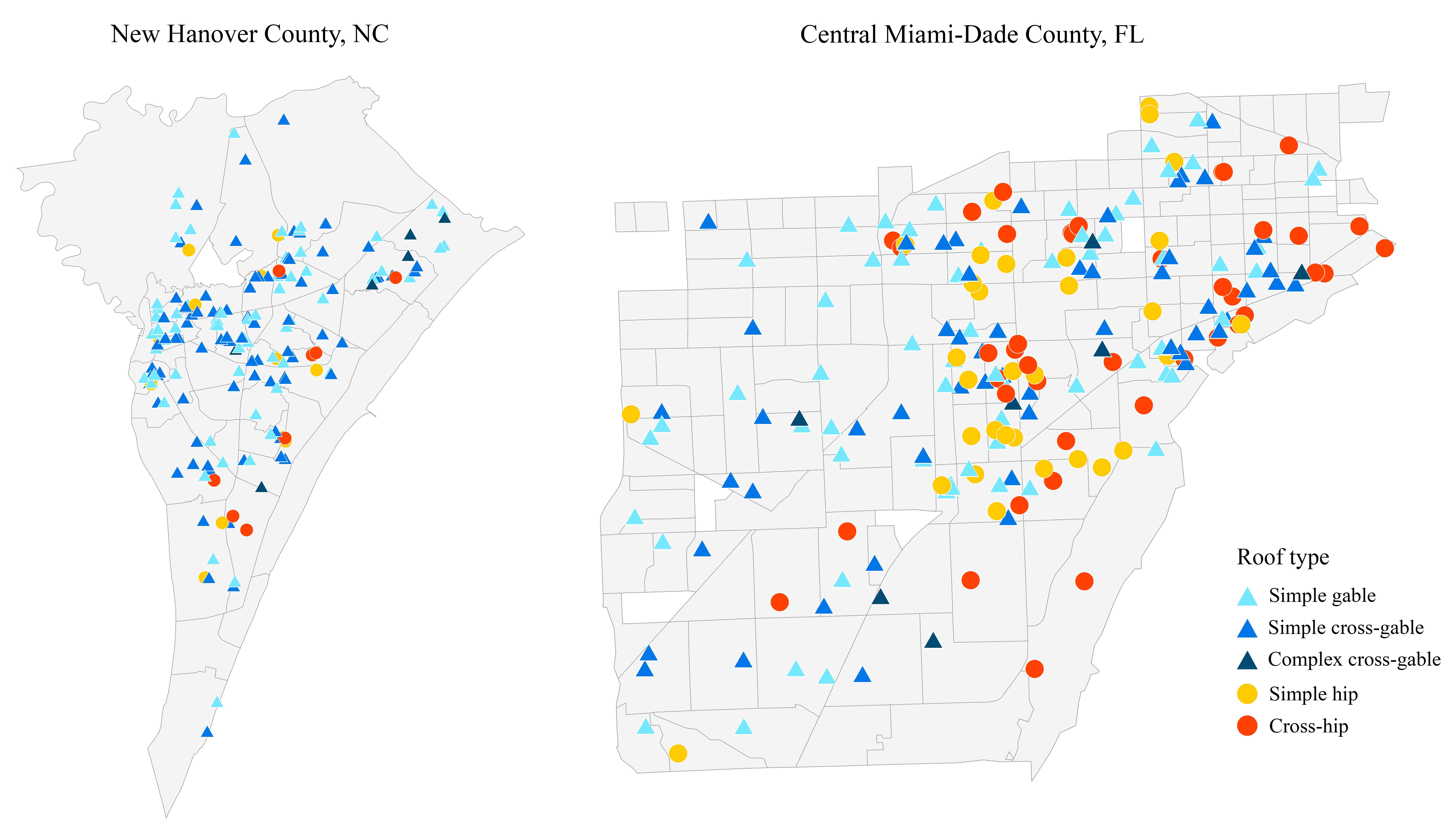}
\caption{Location and ground truth roof type of buildings in the testing dataset.}
\label{test_map}
\end{figure}

Random Forest and Support Vector Machine algorithms were trained to predict roof type and roof complexity for each study area. The 10-fold cross-validation was implemented to assess the performance of each model. Two testing datasets, consisting of 200 houses with unknown roof types for each study area, were created to examine the capability of the models to predict missing roof data. Roof types of the testing buildings were manually labelled based on satellite and street view images downloaded from Google Maps. The locations and the ground truth of roof types of the testing buildings are illustrated in Fig.~\ref{test_map}. The validation and testing accuracy of each model predicting roof type and complexity are listed in Tables~\ref{table:imputation type} and~\ref{table:imputation comp}. Random Forest and Support Vector Machine show similar validation accuracy, while Random Forest performs better on the testing datasets. Moreover, the models achieved higher accuracy in predicting roof type than in predicting roof complexity. As shown in Tables~\ref{table:imputation type} and~\ref{table:imputation comp}, the testing accuracies are generally lower than the validation accuracies, which might be caused by the low-quality neighbor data for the buildings in the testing datasets. Buildings with unknown roof types are usually located in neighborhoods with high vegetation coverage, which increases the probability of those buildings having tree-blocked neighbors. In summary, the imputation models could provide reasonable estimation for the missing roof data, while the performance of the models drops in neighborhoods with high plant density. The prediction of the roof complexity is more sensitive to the quality of neighboring roof data than the prediction of roof type to be gable or hip.   

\begin{table}[H]
\caption{Validation and testing accuracy of predicting roof type (gable or hip) with different models.}
\label{table:imputation type}
\centering
\begin{tabular}{l c c c c}
\hline
\multirow{2}{*}{\makecell{\\Model}}&
\multicolumn{2}{c}{New Hanover} &
\multicolumn{2}{c}{Miami-Dade}\\ \cline{2-5}
& \makecell[tc]{10-fold Cross\\Validation} & \makecell[tc]{Testing\\accuracy} & \makecell[tc]{10-fold Cross\\Validation} & \makecell[tc]{Testing\\accuracy}\\
\hline
Random Forest & 0.85 & 0.89 & 0.78 & 0.72\\
Support Vector Machine & 0.85 & 0.88 & 0.77 & 0.68\\
\hline
\end{tabular}
\end{table}

\begin{table}[H]
\caption{Validation and testing accuracy of predicting roof complexity (simple or complex) with different models.}
\label{table:imputation comp}
\centering
\begin{tabular}{l c c c c}
\hline
\multirow{2}{*}{\makecell{\\Model}}&
\multicolumn{2}{c}{New Hanover} &
\multicolumn{2}{c}{Miami-Dade}\\ \cline{2-5}
& \makecell[tc]{10-fold Cross\\Validation} & \makecell[tc]{Testing\\accuracy} & \makecell[tc]{10-fold Cross\\Validation} & \makecell[tc]{Testing\\accuracy}\\
\hline
Random Forest & 0.81 & 0.64 & 0.77 & 0.66\\
Support Vector Machine & 0.82 & 0.64 & 0.77 & 0.64\\
\hline
\end{tabular}
\end{table}

\section{Discussion}

\noindent The proposed automatic roof classification workflow can be easily applied to other locations where satellite images are available. As alternatives of the ZTRAX database~\cite{zillow2018}, building geolocation can also be obtained from publicly available building footprint databases (\eg OpenStreetMap~\cite{haklay2008}), and critical building characteristics can be found in tax appraisers' databases. The image collection approach aided by the Google APIs directly generates single-building-level satellite images without an additional building detection process~\cite{pi2020convolutional,zhou2018automated}, which largely simplifies the workflow and enhances the quality of images. The application of the CNN model on study areas shows that valid roof type data can be extracted from satellite images for more than 80\% of the houses despite the deficiencies of images (\eg tree occlusion). It is worth noting that the flat roof is excluded from the datasets used for training and testing in this paper since the flat roof is rarely used for single-family residential buildings and is not modeled in existing wind vulnerability models. In satellite images downloaded for study areas, flat roof is not observed in New Hanover County. But very few houses in Miami-Dade County are found to have flat roofs, and most flat roofs were classified as unknown in the roof type dataset generated using the CNN model. Therefore, the actual proportion of images with a clear view of rooftops in Miami-Dade County is higher than the result shown in Section~\ref{data}. 

The CNN model shows the capability of generating roof-type data for large-scale building stocks efficiently and accurately. The detailed roof type datasets for study areas provide quantifiable descriptions of the spatial distribution of roof types at different levels. High variance in roof data statistics among census tracts reveals the great significance of the quality of roof type data to the accuracy of the wind risk assessment. High-resolution building inventory generated with the proposed methods can also support more detailed and refined analysis. Moreover, the distribution of roofs with different complexity levels can facilitate refining building archetypes to provide a more realistic representation of building stock. 

Imputation algorithms were developed to populate roof-type data that cannot be extracted from the images. However, the quality of input features (\ie neighbor type and neighbor complexity) is lower for the buildings with missing roof data than for other buildings, which deteriorates the performance of the models when applied to populate missing data. To improve the prediction accuracy, other data sources (\eg street view images) can be integrated to provide supplementary information for the roof features.  

\section{Conclusions}

\noindent An automated framework is developed to produce high-resolution roof-type datasets for regional wind risk assessment with high accuracy and efficiency. Five roof types are defined based on their wind performance. The roof types, which included non-rectangular roofs, can provide more realistic descriptions of the building stock than the roof types used by existing wind vulnerability models (\eg HAZUS-MH~\cite{vickery2006}, and FPHLM~\cite{pinelli2011}). The pre-trained VGG-19 model was trained to classify roofs using building-level satellite images, and high accuracies were achieved for all classes. The proposed framework was also proved to be capable of identifying dominant roof styles for buildings with mixed roof types. Roof types for 161,772 single-family houses in two study areas were classified using the developed CNN model. Based on city-level roof type distribution, 70\% and 67\% of houses in the study areas had gable and non-rectangular roofs, respectively. Large variances were observed within the census tract-level roof data. The proportion of houses with gable roofs ranged from 1\% to 100\%, and houses with non-rectangular roofs varied from 16\% to 97\% between census tracts. 

To complement the image-based roof classification, imputation algorithms were trained to populate missing roof data with critical building characteristics and roof styles of the neighboring buildings. The roof type of the neighboring buildings plays the most important role in predicting missing roof data. The \textit{year built} and the \textit{building area} metadata were also relevant to the roof type. Random Forest and Support Vector Machine algorithms were trained and evaluated for predicting missing roof data, with Random Forest performing better on testing datasets. The overall accuracies of 89\% and 72\% were achieved for predicting roof type to be gable or hip for the test buildings in New Hanover and Miami-Dade counties, respectively.

The high-resolution roof-type database developed with the proposed workflow provides guidance for developing building archetypes for wind damage modeling. It facilitates the creation of accurate and realistic building inventories for regional wind risk assessment and supports detailed risk analysis beyond the basic census tract level (\eg neighborhood level). The proposed roof classification framework can also be adapted for collecting other building characteristics and applications (\eg solar potential estimation). 

\section{Acknowledgments}
\noindent The authors are grateful for financial support from the National Science Foundation (Awards \#2209190). Some of the building metadata used in this study were provided by Zillow through the Zillow Transaction and Assessment Dataset (ZTRAX). More information on accessing the data can be found at http://www.zillow.com/ztrax. The results and opinions are those of the author(s) and do not reflect the position of NSF or the Zillow Group.

\pagebreak
\bibliography{References}

\begin{thebibliography}{10}
\expandafter\ifx\csname url\endcsname\relax
  \def\url#1{\texttt{#1}}\fi
\expandafter\ifx\csname urlprefix\endcsname\relax\def\urlprefix{URL }\fi
\expandafter\ifx\csname href\endcsname\relax
  \def\href#1#2{#2} \def\path#1{#1}\fi

\bibitem{pita2008}
G.~Pita, J.~Pinelli, C.~Subramanian, K.~Gurley, S.~Hamid, Hurricane
  vulnerability of multi-story residential buildings in florida, Proceedings
  ESREL 2008 (2008).

\bibitem{model2005}
{Florida Public Hurricane Loss Model}, Engineering team final report, vol. i,
  ii, and iii, florida international university (2005).

\bibitem{xu1998}
Y.~L. Xu, G.~Reardon, Variations of wind pressure on hip roofs with roof pitch,
  Journal of Wind Engineering and Industrial Aerodynamics 73~(3) (1998)
  267--284.

\bibitem{gavanski2013}
E.~Gavanski, B.~Kordi, G.~A. Kopp, P.~J. Vickery, Wind loads on roof sheathing
  of houses, Journal of Wind Engineering and Industrial Aerodynamics 114 (2013)
  106--121.

\bibitem{shao2018}
S.~Shao, T.~Stathopoulos, Q.~Yang, Y.~Tian, {Wind pressures on 4$:$12-sloped
  hip roofs of L-and T-shaped low-rise buildings}, Journal of Structural
  Engineering 144~(7) (2018).

\bibitem{Crandell1993}
J.~H. Crandell, M.~Nowak, E.~M. Laatsch, A.~van Overeem, C.~Barbour, R.~Dewey,
  H.~Reigel, H.~Angleton, {Assessment of damage to single-family homes caused
  by Hurricanes Andrew and Iniki}, US Department of Housing and Urban
  Development's Office of Policy Development and Research, 1993.

\bibitem{Brown-Giammanco2018}
T.~M. Brown-Giammanco, I.~M. Giammanco, H.~Pogorzelski, {Hurricane Harvey Wind
  Damage Investigation}, Insurance Institute for Business and Home Safety,
  2018.

\bibitem{pita2011}
G.~Pita, R.~Francis, Z.~Liu, J.~Mitrani-Reiser, S.~Guikema, J.-P. Pinelli,
  Statistical tools for populating/predicting input data of risk analysis
  models, in: Vulnerability, Uncertainty, and Risk: Analysis, Modeling, and
  Management, 2011, pp. 468--476.

\bibitem{mohajeri2018}
N.~Mohajeri, D.~Assouline, B.~Guiboud, A.~Bill, A.~Gudmundsson, J.-L.
  Scartezzini, A city-scale roof shape classification using machine learning
  for solar energy applications, Renewable Energy 121 (2018) 81--93.

\bibitem{bengio2013}
Y.~Bengio, A.~Courville, P.~Vincent, Representation learning: A review and new
  perspectives, IEEE transactions on pattern analysis and machine intelligence
  35~(8) (2013) 1798--1828.

\bibitem{HUTHWOHL2018150}
P.~Hüthwohl, I.~Brilakis,
  \href{https://www.sciencedirect.com/science/article/pii/S1474034617305323}{Detecting
  healthy concrete surfaces}, Advanced Engineering Informatics 37 (2018)
  150--162.
\newblock \href {https://doi.org/https://doi.org/10.1016/j.aei.2018.05.004}
  {\path{doi:https://doi.org/10.1016/j.aei.2018.05.004}}.
\newline\urlprefix\url{https://www.sciencedirect.com/science/article/pii/S1474034617305323}

\bibitem{el2019}
H.~El-Hariri, K.~Mulpuri, A.~Hodgson, R.~Garbi, Comparative evaluation of
  hand-engineered and deep-learned features for neonatal hip bone segmentation
  in ultrasound, in: Medical Image Computing and Computer Assisted
  Intervention--MICCAI 2019: 22nd International Conference, Shenzhen, China,
  October 13--17, 2019, Proceedings, Part II 22, Springer, 2019, pp. 12--20.

\bibitem{alidoost2018}
F.~Alidoost, H.~Arefi, A cnn-based approach for automatic building detection
  and recognition of roof types using a single aerial image, PFG--Journal of
  Photogrammetry, Remote Sensing and Geoinformation Science 86~(5) (2018)
  235--248.

\bibitem{buyukdemircioglu2021}
M.~Buyukdemircioglu, R.~Can, S.~Kocaman, Deep learning based roof type
  classification using very high resolution aerial imagery, The International
  Archives of Photogrammetry, Remote Sensing and Spatial Information Sciences
  43 (2021) 55--60.

\bibitem{wang2021}
C.~Wang, Q.~Yu, K.~H. Law, F.~McKenna, X.~Y. Stella, E.~Taciroglu,
  A.~Zsarn{\'o}czay, W.~Elhaddad, B.~Cetiner, Machine learning-based regional
  scale intelligent modeling of building information for natural hazard risk
  management, Automation in Construction 122 (2021) 103474.

\bibitem{castagno2018}
J.~Castagno, E.~Atkins, Roof shape classification from lidar and satellite
  image data fusion using supervised learning, Sensors 18~(11) (2018) 3960.

\bibitem{jayaseeli2020}
J.~D. Jayaseeli, D.~Malathi, An efficient automated road region extraction from
  high resolution satellite images using improved cuckoo search with
  multi-level thresholding schema, Procedia Computer Science 167 (2020)
  1161--1170.

\bibitem{zambanini2020}
S.~Zambanini, A.-M. Loghin, N.~Pfeifer, E.~M. Soley, R.~Sablatnig, Detection of
  parking cars in stereo satellite images, Remote Sensing 12~(13) (2020) 2170.

\bibitem{vickery2006}
P.~J. Vickery, P.~F. Skerlj, J.~Lin, L.~A. Twisdale~Jr, M.~A. Young, F.~M.
  Lavelle, Hazus-mh hurricane model methodology. ii: Damage and loss
  estimation, Natural Hazards Review 7~(2) (2006) 94--103.

\bibitem{pinelli2011}
J.-P. Pinelli, G.~Pita, K.~Gurley, B.~Torkian, S.~Hamid, C.~Subramanian, Damage
  characterization: Application to florida public hurricane loss model, Natural
  Hazards Review 12~(4) (2011) 190--195.

\bibitem{parackal2016}
K.~Parackal, M.~Humphreys, J.~Ginger, D.~Henderson, Wind loads on contemporary
  australian housing, Australian Journal of Structural Engineering 17~(2)
  (2016) 136--150.

\bibitem{uematsu2022}
Y.~Uematsu, T.~Yambe, A.~Yamamoto, Wind loading of photovoltaic panels
  installed on hip roofs of rectangular and l-shaped low-rise buildings, Wind
  2~(2) (2022) 288--304.

\bibitem{sarma2023}
H.~D. Sarma, I.~Zisis, M.~Matus, Effect of roof shape on wind vulnerability of
  roof sheathing panels, Structural Safety 100 (2023) 102283.

\bibitem{masoomi2018}
H.~Masoomi, M.~R. Ameri, J.~W. van~de Lindt, Wind performance enhancement
  strategies for residential wood-frame buildings, Journal of Performance of
  Constructed Facilities 32~(3) (2018) 04018024.

\bibitem{stewart2018}
M.~G. Stewart, J.~D. Ginger, D.~J. Henderson, P.~C. Ryan, Fragility and climate
  impact assessment of contemporary housing roof sheeting failure due to
  extreme wind, Engineering Structures 171 (2018) 464--475.

\bibitem{zillow2018}
{Zillow}, \href{http://www.zillow.com/ztrax/}{{ZTRAX}: {Zillow} {Transaction}
  and {Assessor} {Dataset}, 2018-{Q2}} (2018).
\newline\urlprefix\url{http://www.zillow.com/ztrax/}

\bibitem{Peng2013}
J.~Peng, Modeling {Natural} {Disaster} {Risk} {Management}: {Integrating} the
  roles of {Insurance} and {Retrofit} and {Multiple} {Stakeholder}
  {Perspectives}, Ph.{D}. {Dissertation}, University of Delaware, Newark,
  Delaware, USA (2013).

\bibitem{meng2023effects}
S.~Meng, C.~J. Williams, R.~A. Davidson, E.~Taciroglu, Effects of roof shape
  and roof pitch on extreme wind fragility for roof sheathing, Journal of
  Structural Engineering 149~(7) (2023) 04023093.

\bibitem{partovi2017}
T.~Partovi, F.~Fraundorfer, S.~Azimi, D.~Marmanis, P.~Reinartz, Roof type
  selection based on patch-based classsification using deep learning for high
  resolution satellite imagery, International Archives of the Photogrammetry,
  Remote Sensing and Spatial Information Sciences-ISPRS Archives 42~(W1) (2017)
  653--657.

\bibitem{lingfors2017}
D.~Lingfors, J.~M. Bright, N.~A. Engerer, J.~Ahlberg, S.~Killinger,
  J.~Wid{\'e}n, Comparing the capability of low-and high-resolution lidar data
  with application to solar resource assessment, roof type classification and
  shading analysis, Applied Energy 205 (2017) 1216--1230.

\bibitem{schwendicke2021}
F.~Schwendicke, A.~Chaurasia, L.~Arsiwala, J.-H. Lee, K.~Elhennawy, P.-G.
  Jost-Brinkmann, F.~Demarco, J.~Krois, Deep learning for cephalometric
  landmark detection: systematic review and meta-analysis, Clinical oral
  investigations 25~(7) (2021) 4299--4309.

\bibitem{zhang2020}
J.~Zhang, Z.~Yin, P.~Chen, S.~Nichele, Emotion recognition using multi-modal
  data and machine learning techniques: A tutorial and review, Information
  Fusion 59 (2020) 103--126.

\bibitem{BERGAMINI2020101052}
L.~Bergamini, M.~Sposato, M.~Pellicciari, M.~Peruzzini, S.~Calderara,
  J.~Schmidt,
  \href{https://www.sciencedirect.com/science/article/pii/S1474034620300215}{Deep
  learning-based method for vision-guided robotic grasping of unknown objects},
  Advanced Engineering Informatics 44 (2020) 101052.
\newblock \href {https://doi.org/https://doi.org/10.1016/j.aei.2020.101052}
  {\path{doi:https://doi.org/10.1016/j.aei.2020.101052}}.
\newline\urlprefix\url{https://www.sciencedirect.com/science/article/pii/S1474034620300215}

\bibitem{simonyan2014}
K.~Simonyan, A.~Zisserman, Very deep convolutional networks for large-scale
  image recognition, arXiv preprint arXiv:1409.1556 (2014).

\bibitem{Russakovsky2015}
O.~Russakovsky, J.~Deng, H.~Su, J.~Krause, S.~Satheesh, S.~Ma, Z.~Huang,
  A.~Karpathy, A.~Khosla, M.~Bernstein, A.~C. Berg,
  \href{http://dx.doi.org/10.1007/s11263-015-0816-y}{{ImageNet Large Scale
  Visual Recognition Challenge}}, International Journal of Computer Vision
  (2015) 211--252\href {https://doi.org/10.1007/s11263-015-0816-y}
  {\path{doi:10.1007/s11263-015-0816-y}}.
\newline\urlprefix\url{http://dx.doi.org/10.1007/s11263-015-0816-y}

\bibitem{Kang2018}
J.~Kang, M.~K{\"{o}}rner, Y.~Wang, H.~Taubenb{\"{o}}ck, X.~Xiang,
  \href{https://doi.org/10.1016/j.isprsjprs.2018.02.006}{{Building instance
  classification using street view images}}, ISPRS Journal of Photogrammetry
  and Remote Sensing 145 (2018) 44--59.
\newblock \href {https://doi.org/10.1016/j.isprsjprs.2018.02.006}
  {\path{doi:10.1016/j.isprsjprs.2018.02.006}}.
\newline\urlprefix\url{https://doi.org/10.1016/j.isprsjprs.2018.02.006}

\bibitem{Hamid2021}
S.~S. Hamid, Florida Public Hurricane Loss Model, version 8.1, ,submitted to
  the Florida Commission on Hurricane Loss Projection Methodology, May 24,
  2021,
  \url{https://fphlm.cs.fiu.edu/files/wind_certification/v8.0Submission/Submission_Document/20210609_FPHLM_2019_FinalizedMar2021_Submission_Document_No_Track_Changes.pdf},
  2021.

\bibitem{taghinezhad2020}
A.~Taghinezhad, C.~J. Friedland, R.~V. Rohli, B.~D. Marx, An imputation of
  first-floor elevation data for the avoided loss analysis of flood-mitigated
  single-family homes in louisiana, united states, Frontiers in Built
  Environment 6 (2020) 138.

\bibitem{soares2021}
D.~d.~B. Soares, F.~Andrieux, B.~Hell, J.~Lenhardt, J.~Badosa, S.~Gavoille,
  S.~Gaiffas, E.~Bacry, Predicting the solar potential of rooftops using image
  segmentation and structured data, arXiv preprint arXiv:2106.15268 (2021).

\bibitem{haklay2008}
M.~Haklay, P.~Weber, Openstreetmap: User-generated street maps, IEEE Pervasive
  computing 7~(4) (2008) 12--18.

\bibitem{pi2020convolutional}
Y.~Pi, N.~D. Nath, A.~H. Behzadan, Convolutional neural networks for object
  detection in aerial imagery for disaster response and recovery, Advanced
  Engineering Informatics 43 (2020) 101009.

\bibitem{zhou2018automated}
Z.~Zhou, J.~Gong, Automated residential building detection from airborne lidar
  data with deep neural networks, Advanced engineering informatics 36 (2018)
  229--241.

\end{thebibliography}

\end{document}